\documentclass[preprint,12pt]{elsarticle}

\usepackage[T1]{fontenc}
\usepackage{amssymb}
\usepackage{amsmath}
\usepackage{multirow}
\usepackage{booktabs}
\usepackage{subcaption}
\usepackage{float}
\usepackage{hyperref}
\usepackage{lineno}

\journal{Ecotoxicology and Environmental Safety}

\newcommand{\q}[1]{``#1''}

\begin{document}

\begin{frontmatter}



\title{Evaluating machine learning models for predicting pesticide toxicity to honey bees}

\author[labelAGH]{Jakub Adamczyk\corref{cor1}\fnref{eq,label1}}
\ead{jadamczy@agh.edu.pl}

\author[labelIBB]{Jakub Poziemski\fnref{eq,label2}}

\author[labelIBB]{Pawel Siedlecki\fnref{label3}}

\affiliation[labelAGH]{organization={Faculty of Computer Science, AGH University of Krakow},
            city={Cracow},
            country={Poland}}
\affiliation[labelIBB]{organization={Institute of Biochemistry and Biophysics of the Polish Academy of Sciences},
            city={Warsaw},
            country={Poland}}

\cortext[cor1]{Corresponding author}

\fntext[eq]{These authors contributed equally to this work}
\fntext[label1]{ORCID 0000-0003-4336-4288}
\fntext[label2]{ORCID 0000-0002-1033-8529}
\fntext[label3]{ORCID 0000-0002-7482-1341}

\begin{abstract}
Small molecules play a critical role in the biomedical, environmental, and agrochemical domains, each with distinct physicochemical requirements and success criteria. Although biomedical research benefits from extensive datasets and established benchmarks, agrochemical data remain scarce, particularly with respect to species-specific toxicity. This work focuses on ApisTox, the most comprehensive dataset of experimentally validated chemical toxicity to the honey bee (\textit{Apis mellifera}), an ecologically vital pollinator. The primary goal of this study was to determine the suitability of diverse machine learning approaches for modeling such toxicity, including molecular fingerprints, graph kernels, and graph neural networks, as well as pretrained models. Comparative analysis with medicinal datasets from the MoleculeNet benchmark reveals that ApisTox represents a distinct chemical space. Performance degradation on non-medicinal datasets, such as \mbox{ApisTox}, demonstrates their limited generalizability of current state-of-the-art algorithms trained solely on biomedical data. Our study highlights the need for more diverse datasets and for targeted model development geared toward the agrochemical domain.
\end{abstract}


\end{frontmatter}

\section{Introduction}

Low molecular weight compounds, often called small molecules, compounds, or just chemicals, play an important role in various research fields, including material science, biomedical research, and agrochemistry among others. Each field has distinct requirements for such molecules that can help define the chance of their success \cite{small_molecules_drug_discovery, small_molecules_agro}. These features are often related to physicochemical or molecular properties, for example, in biomedical research the Lipinski Rule of 5 is frequently used to assess lead and drug likeness \cite{filters_Lipinski}, in agrochemistry similar physicochemical approaches were proposed for different types of pesticides \cite{filters_Hao,filters_Tice} and other types of crop protection compounds \cite{zhang18}. Similarly, important restraints have been established for compound toxicity, allowing both prediction \cite{toxCSM,proTOX} and legislation considering harmful substances (e.g. REACH).

The agrochemistry field shares similarities with biomedical research, often connected with evolutionary conserved mechanisms of actions or even molecular targets \cite{Drug_discovery_agro_1, Drug_discovery_agro_2}. However, significant differences between plants, insects, fungi, animals, and humans are frequently exploited when designing selective and specific small chemicals and biologics \cite{plants_vs_animals, fungicides_moa}. Despite the abundance of medicinal chemistry datasets and benchmarks, such as MoleculeNet \cite{MoleculeNet} and Therapeutics Data Commons (TDC) \cite{TDC}, curated agrochemical and environmental datasets and benchmarks, ready to use for machine learning algorithms, remain scarce. Large-scale resources like ECOTOX \cite{ECOTOX} require significant data curation efforts, such as structural standardization, deduplication, and merging multiple experimental measurements for each compound into a single label for supervised learning \cite{ApisTox}. This presents a significant challenge for developing robust predictive models. For example, while there are numerous models of human toxicity capable of predicting the off-target effects of known chemicals, there are few analogous models for agrochemicals \cite{EPA_PRA}. This gap is especially pronounced in the prediction of toxicity for environmentally important insects such as bees and specific water organisms, or economically important plants like crops.

Large-scale worldwide use of herbicides and other pesticides can cause toxicities to non-target organisms and environmental degradation. A growing number of studies highlight the need for more selective herbicides targeting plant-specific pathways without affecting aquatic species, microbiota, insects, or mammals. \cite{24D,aquaTox,microTox} Research on agrochemical toxicity has become even more essential with respect to environmentally significant insects like honey bee (\textit{Apis mellifera}), which greatly influence both agriculture health and economies. This important pollinator, vulnerable to various chemicals, highlights both the necessity for more precise and species-specific toxicity data, and for more robust predictive models to assess and mitigate toxicity risks.

The main scope of this research is the exploration of ApisTox, the most comprehensive dataset on bee toxicity \cite{ApisTox} with various machine learning approaches to establish current predictive capabilities in this important domain. ApisTox includes chemical compounds represented as SMILES strings, along with their experimentally determined toxicity data for \textit{Apis mellifera}. The potential of ApisTox for the development of predictive methods is assessed, ranging from molecular fingerprints to graph kernels, to graph neural networks, and pretrained neural models. Next, we analyze this dataset in terms of its uniqueness and similarities with respect to selected medicinal datasets included in the MoleculeNet benchmark. Finally, we draw conclusions on what works, what fails, and what types of deficiencies need to be addressed in order for the field of bee toxicity prediction to move forward. Limits of current state-of-the-art machine learning (ML) methods are explored, including comparisons with established methods such as molecular fingerprints or graph kernels. Explainable AI tools are also employed for behavioral testing, to verify the chemical soundness of predictions. As such, this work provides value not only to the ML community but also to experimental researchers and entities involved in toxicity studies, both from scientific and applied perspectives. This study aims to contribute to ongoing efforts to reduce animal testing burdens and to support the development of safer agrochemicals, protect essential pollinators like bees, and promote sustainable agricultural practices.

\section{Materials and methods}

\subsection{ApisTox dataset}

The main data source used in this study is a recently published ApisTox dataset \cite{ApisTox}, which provides toxicity data on pesticides and other agrochemicals for the honey bee (\textit{Apis Mellifera}). ApisTox consists of 1035 compounds, taken from the ECOTOX, PPDB, and BPDB databases, with an applied deduplication and standardization pipeline. This makes it larger and considerably higher quality than previous ML datasets in this area. Notably, it contains a much higher number of 296 toxic molecules, as well as 739 non-toxic ones. The dataset is therefore moderately imbalanced, with 29\% of pesticides being toxic. It is overall quite large, compared to most targeted agrochemical datasets in this area prepared for ML modeling \cite{ApisTox}, e.g. CropCSM \cite{datasets_CropCSM} (871 molecules), BeeTOX \cite{BeeTOX} (876 molecules), or BeeToxAI \cite{datasets_BeeToxAI} (472 molecules). Further, ApisTox contains a very significant proportion of currently used pesticides. As such, it is highly representative of the overall pesticide chemical space while also providing annotations for the bee toxicity target.

Uniquely, ApisTox provides predetermined train-test splits, which ensures a fair comparison of different algorithms: maximum diversity split (MaxMin) \cite{ApisTox}, as well as approximation of time split, based on PubChem literature reference. Those split approaches are shown to be potentially quite challenging for ML algorithms, with diverse test sets and high train-test separation, avoiding data leakage related to molecular similarity. This allows comparing various ML-based molecular property prediction methods and extend this benchmark in the future.

In machine learning terms, the task in this work is binary classification of toxic vs non-toxic molecules, primarily based on their molecular graph representation. The labels in ApisTox denote the most serious toxicity type, i.e. contact, oral, or other. As such, molecules in the positive class are highly toxic in at least one way to honey bees and can be considered unsafe to use.

\subsection{Molecular dataset analysis methods}

The original ApisTox publication \cite{ApisTox} contains many analyses, in particular showing that toxic and non-toxic pesticides vary in terms of molecular structure but not in basic physicochemical properties. In this work, we further analyze the properties of this dataset, showing that it covers a specific agrochemistry-related chemical space \cite{Drug_discovery_agro_1,small_molecules_agro}. This sets it apart from most benchmarks in molecular property prediction, which are focused on medicinal chemistry, e.g. MoleculeNet \cite{MoleculeNet} and Therapeutics Data Commons (TDC) \cite{TDC}. In this section, we describe the chemoinformatics tools used for analysis of ApisTox and inter-dataset differences.

\subsubsection{Molecular filters}

Molecular filters are a set of rules used to select and limit a set of chemical molecules, which leads to obtaining the desired chemical space \cite{molecular_filters}. They are used, among other things, in molecular screening to reduce the number of compounds to be analyzed by eliminating undesirable structures, such as those with low bioavailability or potential toxicity \cite{molecular_filters,molecular_filters_2}. Molecular filters can be divided into two main categories: structural filters, which classify molecules on the basis of the presence of specific functional groups or chemical fragments, and physicochemical filters, which evaluate compounds for selected properties such as molecular weight, logP, or the number of hydrogen bond donors and acceptors. A molecule passes the filter if it meets its criteria, e.g. does not contain any problematic groups, or has the properties in desired ranges. Molecular filter analysis allows the characterization and approximation of the types of chemical structures present in a given set of molecules \cite{molecular_filters}.

Lipinski \cite{filters_Lipinski} and Ghose \cite{filters_Ghose} filters were designed to identify drug-like compounds, focusing on parameters related to adsorption. Their criteria include molecular weight, number of hydrogen bonds, and solubility. Hao filter \cite{filters_Hao} aims to detect molecules with pesticide-specific properties, while the Tice Insecticides filter \cite{filters_Tice} is used to identify potential insecticides, taking into account the specific characteristics of this pesticide type. The Brenk filter \cite{filters_Brenk} was developed for high-throughput screening (HTS). Specifically, it focuses on the elimination of compounds containing problematic functional groups that can exhibit high toxicity or adverse pharmacokinetic properties.

\subsubsection{Molecular diversity}

The chemical space diversity can directly influence the level of difficulty in performing the classification. It affects transfer learning performance, as the discordance between pretraining and downstream data affects the effectiveness of the model's prior knowledge. It is also highly relevant from an ML benchmarking perspective, as datasets should ideally be orthogonal and cover unique chemical spaces to verify the model performance in varying conditions.

One of the simplest methods of analyzing such properties is comparing the distribution or types of chemical elements present in molecules, as well as substructures, functional groups, or fragments. A slight problem with such an analysis is that there is no single commonly accepted definition of a functional group or chemical fragment in the literature. Different approaches are usually used, based on different sets of functional groups, typically defined and published as collections of SMARTS patterns. Such methods are also often used as molecular fingerprints for building QSAR predictive models. Examples include the Laggner fingerprint and MACCS Keys.

Arguably, the most popular method of comparing chemical spaces of datasets is checking the average pairwise Tanimoto of similarity between their molecules. Molecular fingerprints are typically used as a vectorization method, e.g. ECFP4 hashed fingerprint, or substructural PubChem fingerprint. The Tanimoto coefficient is the most widely used similarity measure for comparison of resulting fingerprint vectors.

ECFP4 fingerprint \cite{fp_ecfp} is most commonly used as a vectorization method. It is an intuitive measure, with interpretable values. However, this approach has some downsides, such as being susceptible to clumping (clustering) of molecules, and is often only modestly correlated with biological activity \cite{diversity_ncircles}.

\#Circles measure \cite{diversity_ncircles} has been proposed as a solution to these problems. Inspired by sphere exclusion clustering, it measures the coverage of the chemical space by identifying the local neighborhoods occupied by a given set of molecules. It counts the maximum number of mutually exclusive circles (hyperspheres) with a given radius in a dataset. Formally, given a set $ \mathcal{S} $ with $ n $ molecules, distance measure $ d $ and threshold $ t \in \left[0, 1 \right) $, we define \#Circles measure as:
\begin{equation}
\#Circles(\mathcal{S}, d, t) = \max_{\mathcal{C}\subseteq\mathcal{S}} \left| \mathcal{C} \right |\quad \text{where} \quad d(x,y) \ge t \quad  \forall x \neq y \in \mathcal{C} 
\end{equation}

\subsection{Molecular property prediction}

The task of predicting the toxicity of pesticides here is an example of molecular property prediction, also known as quantitative structure-property prediction (QSPR). It is typically modeled as graph classification, where the input molecule is processed as an undirected attributed graph. Various approaches have been proposed, ranging from manual feature engineering, automated feature extraction or similarity measurement algorithms combined with Random Forrest (RF) or XGB models, to graph representation learning, exemplified by graph neural networks (GNNs). Recently, pretrained models for molecules have also been proposed, based on either GNNs or text-based SMILES representation. We describe them in detail in the following subsections.

\subsubsection{Molecular fingerprints}

Molecular fingerprints are a group of related methods for vectorizing molecules. They extract feature vectors in various ways, depending on the type of fingerprint. Arguably, they are the most commonly used tool for QSPR in chemoinformatics, explaining the existing multitude of available algorithms.

Descriptors, or descriptor sets, such as Mordred \cite{fp_mordred}, compute a large, predefined set of molecular features, e.g. distribution of element types, topological indexes (e.g. Wiener, Zagreb), or substructure counts (e.g. rings, paths). They are highly interpretable and can achieve high performance when the target property is well correlated with physicochemical or topological features of the molecule.

Substructure fingerprints extract a predefined set of subgraphs, e.g. functional groups, aromatic rings, or atoms of a given element. The choice of substructures is typically selected by domain experts, most typically medicinal chemists. One can check the existence of a given substructure (binary fingerprint) or count its occurrences in a molecule (count fingerprint). Examples include MACCS Keys and PubChem fingerprint. They are very interpretable and can encode domain-relevant features.

Hashed fingerprints define the general shape of subgraphs that are extracted from a molecule. Those can be, for example, circular neighborhoods for the ECFP fingerprint or shortest paths between pairs of atoms in the Atom Pair fingerprint . A unique identifier is assigned to each subgraph, which is then hashed onto a constant-length output vector at a position based on the identifier value. In binary fingerprints, we ignore hashing collisions, whereas count fingerprints sum up all occurrences at a given index. They are the most flexible group, often resulting in very strong ML classifiers, competitive with complex GNNs \cite{scikit_fingerprints}.

\subsubsection{Feature engineering baselines}

Multiple works have shown the surprising efficiency of simple baseline algorithms, compared to GNNs. Therefore, we include them to fairly assess the performance of more sophisticated methods.

In one of the first papers on fair evaluation of GNNs for graph classification \cite{fair_comparison_GNNs}, the authors proposed to simply count atoms of different elements. This approach ignores the molecular graph topology, relying only on the distribution of the simplest atom attribute. However, it shows competitive performance on bioinformatics problems.

On the other hand, Local Topological Profile (LTP) \cite{LTP} baseline uses graph topology exclusively. It creates a feature vector for a graph, based on concatenated histograms of topological descriptors: vertex degree, statistics of neighbors (bonded atoms) degrees (min, max, average, standard deviation), edge betweenness centrality (EBC), Jaccard Index (JI), and Local Degree Score (LDS). This approach aims to capture both local and global properties of graphs and is shown to be very competitive with complex GNNs.

Molecular Topological Profile (MOLTOP) \cite{MOLTOP} proposed a unified approach, with topological and molecular features, aimed specifically at molecular property prediction. First, histograms of topological descriptors are used, similarly to LTP, but different descriptors are selected, based on their chemoinformatical interpretation: EBC, Adjusted Rand Index (ARI), and SCAN Structural Similarity score. Then, simple statistics of basic molecular features are added to the feature vector: average, sum, and standard deviation of atom elements and bond orders. This fusion of two information sources is shown to give a particularly strong baseline for QSPR.

\subsubsection{Graph kernels}

Graph kernels quantify the similarity between graphs, based on their structure \cite{graph_kernels}. In many cases, they also allow for incorporating vertex attributes, such as element type, and many molecule kernels have been proposed to describe the structural similarity of compounds. The resulting kernel matrix, containing pairwise similarity values between molecules, allows for the use of any established kernel methods, such as the SVM classifier. Due to their high expressive power, they show excellent performance in chemoinformatics \cite{graph_kernels_cheminformatics}.

The simplest possible kernels are the vertex and edge histogram kernels, which use a dot product between the counts of elements and the bonds between compounds, respectively. Many more complex kernels have been proposed, capturing local and global topology of graphs, e.g. shortest paths \cite{graph_kernel_shortest_paths} or graphlets (all subgraphs up to a given size) \cite{graph_kernel_graphlet}. The propagation kernel \cite{graph_kernel_propagation} has been proposed to diffuse the node information in the graph, strongly focusing on the functional similarity between molecules, in addition to their topology.

The Weisfeiler-Lehman (WL) kernel \cite{graph_kernel_wl} is very commonly used, due to its excellent performance, speed, and strong theoretical properties. It is based on the Weisfeiler-Lehman isomorphism test \cite{WL_test}, aimed at distinguishing non-isomorphic graphs. The idea is to exchange information between neighboring nodes (bonded atoms), which captures the topological structure of a $k$-hop neighborhood around each atom after $k$ iterations. In particular, it has also directly influenced the Morgan algorithm for the molecular graph isomorphism test and the ECFP fingerprint. This isomorphism testing framework also underlies the theory behind message-passing GNNs. The WL kernel is particularly computationally efficient, compared to most graph kernels.

The Weisfeiler-Lehman Optimal Assignment (WL-OA) kernel \cite{graph_kernel_wl_oa} has been proposed as an extension. The idea is that two similar molecules should have matching parts, e.g. same or similar rings or functional groups. This is realized by the optimal assignment kernel, which acts as a function mapping atoms in two molecules (identified using the Weisfeiler-Lehman procedure) to maximize the sum of the pairwise atom similarities. This substantially strengthens the ability of the kernel to distinguish between similar and dissimilar molecular substructures. Thus, it achieves very strong classification performance, albeit at a considerably higher computational cost.

\subsubsection{Graph neural networks (GNNs)}

Graph neural networks (GNNs) \cite{GNNs_intro} are a family of neural networks for graph representation learning, with appropriate inductive biases and theoretical guarantees. In particular, their operations, like graph convolution, are permutation-invariant, allowing them to process vertices and edges without assuming any natural order. They also naturally incorporate any atom and bond features. This makes them a natural choice for learning embeddings of molecules.

Most modern GNNs utilize various types of graph convolution, with additions mostly to capture long-range relationships or enhance whole graph embedding. They are based on the message-passing paradigm, where in each layer node sends a message to its neighbors (bonded atoms). The message is its current embedding, starting with initial atom features for the input molecule. Self-loops, artificial edges from the vertex to itself, are also frequently added to increase the importance of individual atoms where necessary. Graph convolutions differ by how they update the vertex embedding, based on the received messages. To obtain the molecule graph embedding, the final atom embeddings are combined in the readout layer. This is typically a simple column-wise average, sum, or maximum.

Graph Convolutional Network (GCN) \cite{GCN}, based on regularized spectral graph convolution, takes a mean of the messages from the neighbors and the atom itself, weighting them by their degrees. GraphSAGE \cite{GraphSAGE} has three variants, with ``mean'' variant being the most popular. It is quite similar to GCN, but first takes a simple mean of neighbors' messages, and concatenates it to the vertex embedding before update. This separate treatment improves performance on heterophilous tasks \cite{GraphSAGE_2}, e.g. where the atom itself is much more important than its relationship to the neighborhood. The authors of Graph Isomorphism Network (GIN) \cite{GIN} show that all message-passing GNNs are at most as powerful as the 1-WL test in distinguishing isomorphic graphs. Designed GIN convolution, based on multilayer perceptron (MLP) and sum pooling readout, is proven to be as powerful as WL test, resulting in good performance for graph classification. Graph Attention Network (GAT) \cite{GAT} incorporates the attention mechanism to weight neighbors' messages.

$k$-layer GNN is able to aggregate information about the $k$-hop neighborhood around each atom in its embedding. This is important for molecular classification since, e.g. understanding ring structures requires many layers under this paradigm. However, building deep GNNs has proven difficult, due to oversmoothing, oversquashing, and unstable training \cite{GNN_training}, which make it difficult to capture long-range dependencies. Jumping Knowledge \cite{GNNs_JK} has been proposed as a simple solution, where we add skip connections, utilizing atom embeddings after each layer to obtain the molecule embedding, instead of just vectors after the final layer. Normalization layers, such as LayerNorm can help stabilize training \cite{pretrained_MAT}. Bayesian hyperparameter optimization (HPO), e.g. using Parzen tree estimator \cite{HPO}, can be used to efficiently select good hyperparameter values, like learning rate or number of layers, enhancing the performance of the resulting architecture \cite{AttentiveFP}.

AttentiveFP \cite{AttentiveFP} applied many of these ideas specifically for molecular property prediction. It utilizes the attention mechanism combined with artificial bonds for capturing long-range dependencies. It also uses the GRU recurrent network for graph level readout, to focus on the most relevant atom embeddings. They also utilized Bayesian HPO to tune many hyperparameters of this architecture.

\subsubsection{Pretrained neural networks}

Most molecular property prediction datasets, including ApisTox, are relatively small. Transfer learning uses neural networks trained on massive general datasets, leveraging this previous knowledge for more specific tasks. Such models are typically based either on graph transformers, a particular type of GNN based on self-attention mechanism, or on text transformers operating on SMILES.

Those models can be finetuned, using the pretraining model as a weight initialization \cite{pretrained_MolFormer}. However, this approach assumes that there is enough data available to avoid overfitting the model, which is often not the case for highly specific domains in chemistry. Instead, one can use the molecule embeddings from the pretrained model, like a molecular fingerprint, and input it to any conventional ML model. This approach is equivalent to freezing most model weights and training only the classification head. It is also known as frozen embeddings, and has shown results competitive with finetuning models, while being faster and less prone to overfitting \cite{pretrained_MolFormer}.

Molecule Attention Transformer \cite{pretrained_MAT} and its later extension, the Relative Molecule Self-attention Transformer (R-MAT) \cite{pretrained_RMAT}, incorporate many inductive biases specific to molecular chemistry into the GNN graph transformer and their pre-training process. They are based on self-attention between atoms, which is a weighted sum of three components, incorporating molecule-specific inductive biases. In particular, both utilize key-value attention between atom embeddings and pairwise shortest paths between atoms, incorporating both local and global molecule context. Both follow the encoder-only transformer architecture, interleaving layers of attention, normalization, and MLPs. The learned atom embeddings are combined with the average readout in MAT and with the attention-based pooling in R-MAT. The pretraining procedure for MAT is based only on masked node reconstruction \cite{pretrained_MAT}, which is a graph-based equivalent of masked language modeling. R-MAT uses more challenging masked subgraph reconstruction (similar to GROVER \cite{pretrained_GROVER}), and multitask regression at the molecule level.

GROVER \cite{pretrained_GROVER} is a GNN based on a graph transformer architecture. It creates embeddings for both atoms and bonds in parallel for increased expressivity, and includes skip connections and randomized message passing to better learn long-range relations in molecules. It proposed using two pretraining tasks: masked subgraph reconstruction, and classification of functional groups existence. This incorporates strong subgraph-level knowledge.

A notable family of novel molecular models are transformers based solely on SMILES, as exemplified by ChemBERTa \cite{pretrained_ChemBERTa}. It follows the decoder-only architecture and can be pretrained only on masked language modeling (MLM) on SMILES. Alternatively, it can be pretrained on multitask regression (MTR), predicting many physicochemical properties of molecules computed with RDKit. BERT-like models have relatively simple architectures but result in highly performant embeddings.

Lastly, a hybrid model between graph-based and SMILES-based models is Mol2Vec \cite{mol2vec}. It treats the molecule as a sequence of words, based on ECFP invariants \cite{fp_ecfp}, computed from the molecular graph. Then it vectorizes them like words, using a pretrained Word2Vec model. Those embeddings are then added to obtain the embedding for the whole molecule.

\subsubsection{Explainability}

Explainable artificial intelligence (XAI), also known as interpretable or explainable machine learning, concerns understanding the behavior and reasoning of the model. This helps ensure that the model learns meaningful relations in the data, instead of random noise or unintended statistical patterns in the positive and negative classes.

We focus on a local, model-agnostic interpretability methods, and specifically counterfactual explanations. Those methods interpret a prediction made for a single molecule with an easy visualization of the results. They are independent of the model used, which is very useful as we explore a wide variety of approaches to molecular property prediction. For binary classification, as in ApisTox, counterfactual explanations are very useful, answering the following question: ``What is the minimal change required for the input molecule so that the model changes the prediction to the opposite class?''. Algorithms of this type can suggest substituents, functional groups, or fragment removal from the input molecule, so that the model changes the prediction, e.g. from non-toxic to toxic.

Molecular Model Agnostic Counterfactual Explanations (MMACE) \cite{exmol} has been designed as a fast and easy-to-use method in this group. Importantly, it does not require dataset-specific training, instead relying on directly modifying the molecule being explained using the STONED \cite{STONED} generative model. In this way, it explored the chemical space around it, using only model predictions to determine samples from the opposite class. Tanimoto similarity on ECFP4 fingerprints is used to select counterfactuals most similar to the original molecule, that is, with the smallest change.

\subsection{Experimental setup}

Here, we describe the experimental procedure and implementation details for our experiments.

\subsubsection{Models implementation}

We implement molecular fingerprints using the scikit-fingerprints library \cite{scikit_fingerprints}, using all 2D fingerprints available in the library version 1.12.0. We omit 3D conformer-based fingerprints, since they gave very subpar results in initial experiments. For all fingerprints, we use Random Forest classifier from scikit-learn \cite{scikit_learn}, with 100 trees (default value) and entropy as split criterion. Entropy worked slightly better than the default Gini entropy in our initial experiments, in agreement with previous works \cite{LTP,MOLTOP}. For model regularization, we set the minimal number of samples to split.

Atom counts baseline \cite{fair_comparison_GNNs} is implemented using RDKit, with Random Forest like for fingerprints. For LTP \cite{LTP} and MOLTOP \cite{MOLTOP} baselines, we use the original code and classifier settings. Both use Random Forest, but with default settings tuned over a large number of datasets.

Graph kernels are implemented using GraKeL \cite{GraKeL}, with the kernel SVM classifier from scikit-learn. After initial experiments, we decided to use the following kernels: shortest paths, propagation, WL, WL-OA. Molecular graphs are attributed with atomic numbers and bond types (single, double, triple, aromatic, other). We tune kernel-specific hyperparameters (see \ref{appendix_hyperparameters} for details) and SVM regularization $C$.

GNNs are implemented in PyTorch \cite{PyTorch} and PyTorch Geometric \cite{PyTorchGeometric}. We implement GCN, GraphSAGE, GIN, GAT, and AttentiveFP. Following many architectures in a popular OGB benchmark \cite{OGB}, we extract a standard set of features for atoms and bonds, and use the embedding layer to obtain their initial vectors. Atom features are based on invariants used in ECFP, e.g. atomic number, degree, formal charge. However, we note that only GIN, GAT, and AttentiveFP make use of bond features. We also use Jumping Knowledge (concatenation variant) \cite{GNNs_JK} and LayerNorm normalization \cite{pretrained_MAT}. For readout, we use sum pooling, proven to be the most expressive \cite{GIN}. We make an exception for AttentiveFP, which uses its original GRU-based pooling instead. We use the same number of channels in each layer and treat the number of layers and channels as hyperparameters, while also tuning the learning rate and dropout. Networks are trained for 100 epochs, without early stopping.

For pretrained models, we use MAT, R-MAT, GROVER, ChemBERTa, and Mol2Vec. All are used as embedding models, similar to fingerprints. We do not perform fine-tuning, since during initial experiments it constantly overfitted and gave worse result than the frozen embeddings approach. Initial experiments indicated that logistic regression performs considerably better than Random Forest as a classifier in this case, probably because pretraining used such linear classifiers and the feature space is better aligned for them. Therefore, we use logistic regression as classifier and tune the regularization strength $C$. For further details on those models, see \ref{appendix_pretrained_full_results}.

For all classifiers, we use class weighting, since ApisTox is imbalanced. We use \textit{balanced} setting from scikit-learn, which uses the class weight inverse to its ratio in the dataset. For hyperparameter tuning, in all cases we use stratified 5-fold cross-validation (CV), selecting the model with the highest AUROC (Area Under Receiver Operating Characteristic curve) value.

\subsubsection{Evaluation}

For a realistic and challenging train-test split, we use two splits provided with ApisTox: MaxMin split and time split approximation. The test set in MaxMin split is created by selecting maximally diverse molecules, i.e. with the highest sum of pairwise Tanimoto distances between their ECFP4 fingerprints. This way, we require the toxicity classifier to perform well across the whole chemical space of the dataset. Since the maximum diversity picking problem is NP-hard, MaxMin heuristic from RDKit is used here \cite{ApisTox}. Time split is approximated by PubChem literature data, with the test set consisting of the newest agrochemicals. This requires out-of-distribution generalization capabilities, since new pesticides are typically structurally novel. For both splits, the test set is 20\% of the data.

To validate the performance of the model, we measure multiple metrics, as recommended in ApisTox paper \cite{ApisTox}. We select metrics appropriate for imbalanced datasets: AUROC, precision, recall, and Matthews correlation coefficient (MCC). Comparison of models based on multiple metrics is important, as they highlight potential differences in their performance. In particular, MCC is reported to better reflect model performance compared to AUROC, and precision and recall focus on false positives and false negatives, respectively.

To achieve more robust estimation of test performance, we retrain models based on Random Forest or GNNs with 50 random seeds (after hyperparameter tuning), and report average and standard deviation of metrics on the test set. This was not possible for SVM, which is deterministic, and for pretrained neural networks, as logistic regression used on top of their embeddings always converged to the same result. Thus, for them, we report only a single score.

We include the hyperparameter ranges and other details in the \ref{appendix_hyperparameters}.

\subsection{Molecular dataset analysis}

RDKit \cite{RDKit} was used to calculate molecular descriptors and most chemoinformatic analyzes. scikit-fingerprints \cite{scikit_fingerprints} was used to calculate molecular fingerprints and diversity calculations.

\#Circles algorithm was implemented using the description from the original publication \cite{diversity_ncircles}, as well as the official implementation. As the exact computation has exponential complexity, we use the fast sequential approximation proposed in the original implementation. We use the same parameters as the original publication, i.e. threshold 0.75 of Tanimoto distance and ECFP4 fingerprint of length 1024 as the feature space.

\subsection{Explainability}

We use the original MMACE implementation from the ExMol package \cite{exmol}. We generate 5000 samples and do not use drug-like filtering (used in the original MMACE paper), in order to properly explore the pesticides chemical space. To avoid data leakage, we explain only the model predictions on molecules from the test set.

\section{Data and software availability}

We release our code publicly at \url{https://github.com/j-adamczyk/ApisTox_bee_toxicity_ML_prediction}. The ApisTox dataset, along with MaxMin and time splits, is available on Zenodo (\url{https://doi.org/10.5281/zenodo.13350981}) and GitHub (\url{https://github.com/j-adamczyk/ApisTox_dataset}). We also include it in our GitHub repository with code for reproducing experimental results.

\texttt{uv} was used as a dependency manager, and we also distribute the \texttt{pyproject.toml} and \texttt{uv.lock} files with exact versions of both direct and transitive library dependencies. All libraries used are open source and can be downloaded with \texttt{uv} or directly with \texttt{pip}.

\section{Results and discussion}

The main results and quality metrics are reported in Table \ref{results_maxmin_split} (MaxMin split) and Table \ref{results_time_split} (time split). For brevity, we include the top 5 fingerprints (with the highest MCC scores), with results for the rest in the \ref{appendix_fingerprints_full_results}. The best metric value in each group is marked in bold (two values in case of ties).

The first observation is that the baseline algorithms perform quite strongly, with MOLTOP being the best on both splits. It even obtains results better than almost all GNNs and pretrained neural networks. However, most other methods, particularly fingerprints and graph kernels (WL and WL-OA), significantly beat baseline results. This is a necessary condition for these methods to be useful. We perform a deeper analysis of model failures in further sections.

WL-OA graph kernel obtains very strong results, with the highest MCC and AUROC on MaxMin split. Its performance on time split is also strong but worse than fingerprints. As MinMax tests interpolative generalization performance, inside and across the chemical space, this setting may benefit the pairwise kernel approaches. However, extrapolating to novel chemical spaces under time split is much more challenging, however, for those very powerful similarity measures. This signifies the importance of evaluating algorithms with different splits.

Molecular fingerprints are, overall, the most performant and robust solution. Interestingly, quite different fingerprints are the best for both splits. For MaxMin, the best fingerprint is Avalon, with 3 substructural fingerprints (Laggner, MACCS, PubChem) also among the top 5. However, for time split, all best fingerprints are hashed, e.g. ECFP, Layered, RDKit. This confirms that flexible, more data-driven hashed fingerprints are better suited for out-of-distribution generalization. In contrast, in-domain predictions can be made accurately by extracting common functional groups and other substructural patterns. Since a very large number of fingerprint algorithms are available, understanding such implications can be useful for practical applications.

The performance of GNNs trained from scratch is very underwhelming in both cases. All are unable to outperform baselines, even the simplest atom count in the MaxMin split. This is despite significant computational resources and extensive hyperparameter tuning, even for AttentiveFP, designed specifically for molecular property prediction. However, GNNs are known to require larger amounts of data to train effectively, and by its nature, agrochemistry requires low-data ML. Furthermore, it frequently involves salts, mixtures, and other molecules with disconnected components in their molecular graphs \cite{ApisTox}. This is problematic for message-passing models, since they can obtain this information only in the last readout layer, limiting the effectiveness of node embedding learning from their neighborhoods.

Our initial assumption was that the pretrained neural networks would solve some of the aforementioned problems, thanks to e.g. transfer learning or usage of graph transformers operating on fully connected graphs. However, those models resulted in surprisingly poor performance, not better than GNN, and failed to outperform baselines in most cases. One possible reason for the failure of MAT, R-MAT, and GROVER is that they are pretrained only on heavily filtered, drug-like compounds from ZINC. Agrochemical compounds may lie outside this chemical space, limiting the usefulness of transfer learning. The GROVER model gave the worst overall result on time split, and second worst on MaxMin split. One possible reason are relatively weak inductive biases for molecular chemistry, compared to e.g. Mol2Vec, MAT or R-MAT, since it relies only on substructure detection.

Overall, those results are quite different from those found on MoleculeNet, the main test bench for molecular models. The problem with relying on a small benchmark is that it becomes oversaturated over time, resulting in models ``overtuned'' to achieve relatively minor improvements just on those standard datasets. Another explanation is that medicinal chemistry datasets such as those from MoleculeNet and TDC verify generalization capabilities only on small drug-like compounds, whereas agrochemistry presents a bit different chemical space. In further sections, we perform deeper analyses in this regard.

\begin{table}
\centering
\resizebox{\textwidth}{!}{
\begin{tabular}{|c|c|c|c|c|c|}
\hline
\textbf{Group} & \textbf{Method} & \textbf{MCC} & \textbf{AUROC} & \textbf{Precision} & \textbf{Recall} \\ \hline
\multirow{5}{*}{Fingerprints} & Avalon & \textbf{0.48 $\pm$ 0.03} & 76.09\% $\pm$ 1.27\% & \textbf{76.37\% $\pm$ 4.40\%} & 39.67\% $\pm$ 1.48\% \\ \cline{2-6} 
 & Laggner & 0.46 $\pm$ 0.03 & 77.45\% $\pm$ 0.84\% & 57.40\% $\pm$ 3.16\% & \textbf{56.95\% $\pm$ 3.01\%} \\ \cline{2-6} 
 & AtomPairs & 0.45 $\pm$ 0.03 & 76.65\% $\pm$ 1.25\% & 70.16\% $\pm$ 3.63\% & 39.67\% $\pm$ 2.46\% \\ \cline{2-6} 
 & MACCS & 0.45 $\pm$ 0.03 & \textbf{79.77\% $\pm$ 0.96\%} & 67.30\% $\pm$ 3.31\% & 42.62\% $\pm$ 3.10\% \\ \cline{2-6} 
 & PubChem & 0.44 $\pm$ 0.03 & 77.25\% $\pm$ 0.99\% & 61.32\% $\pm$ 3.65\% & 47.86\% $\pm$ 2.03\% \\ \hline
\multirow{3}{*}{Baselines} & Atom counts & \textbf{0.36 $\pm$ 0.03} & \textbf{81.01\% $\pm$ 0.70\%} & 49.93\% $\pm$ 2.79\% & \textbf{46.43\% $\pm$ 3.27\%} \\ \cline{2-6} 
 & LTP & 0.18 $\pm$ 0.02 & 66.45\% $\pm$ 0.48\% & 40.58\% $\pm$ 2.88\% & 22.14\% $\pm$ 1.98\% \\ \cline{2-6} 
 & MOLTOP & \textbf{0.36 $\pm$ 0.03} & 76.05\% $\pm$ 0.45\% & \textbf{60.22\% $\pm$ 2.84\%} & 34.81\% $\pm$ 2.42\% \\ \hline
\multirow{4}{*}{\begin{tabular}[c]{@{}c@{}}Graph\\ kernels\end{tabular}} & Propagation & 0.32 & 71.41\% & 51.72\% & 35.71\% \\ \cline{2-6} 
 & Shortest paths & 0.29 & 76.33\% & 41.67\% & 47.62\% \\ \cline{2-6} 
 & WL & 0.42 & 78.47\% & 59.38\% & 45.24\% \\ \cline{2-6} 
 & WL-OA & \textbf{0.49} & \textbf{83.95\%} & \textbf{62.16\%} & \textbf{54.76\%} \\ \hline
\multirow{5}{*}{GNNs} & GCN & 0.25 $\pm$ 0.04 & 71.59\% $\pm$ 1.07\% & 36.16\% $\pm$ 3.18\% & 50.67\% $\pm$ 7.56\% \\ \cline{2-6} 
 & GraphSAGE & 0.31 $\pm$ 0.05 & 71.78\% $\pm$ 1.85\% & \textbf{44.22\% $\pm$ 6.65\%} & 48.33\% $\pm$ 9.16\% \\ \cline{2-6} 
 & GIN & 0.24 $\pm$ 0.04 & 69.37\% $\pm$ 1.61\% & 34.03\% $\pm$ 3.38\% & 56.24\% $\pm$ 11.17\% \\ \cline{2-6} 
 & GAT & 0.26 $\pm$ 0.03 & 71.83\% $\pm$ 1.52\% & 37.18\% $\pm$ 4.67\% & 53.19\% $\pm$ 6.61\% \\ \cline{2-6} 
 & AttentiveFP & \textbf{0.35 $\pm$ 0.04} & \textbf{75.20\% $\pm$ 1.92\%} & 42.54\% $\pm$ 3.39\% & \textbf{60.29\% $\pm$ 4.14\%} \\ \hline
\multirow{5}{*}{\begin{tabular}[c]{@{}c@{}}Pretrained\\ neural\\ networks\end{tabular}} & MAT & 0.36 & 72.29\% & 40.58\% & \textbf{66.67\%} \\ \cline{2-6} 
 & R-MAT & 0.31 & 70.46\% & 36.99\% & 64.29\% \\ \cline{2-6} 
 & GROVER & 0.22 & 71.46\% & 32.86\% & 54.76\% \\ \cline{2-6} 
 & ChemBERTa & \textbf{0.37} & 74.46\% & \textbf{42.86\%} & 64.29\% \\ \cline{2-6} 
 & Mol2Vec & 0.34 & \textbf{76.57\%} & 40.62\% & 61.90\% \\ \hline
\end{tabular}
}
\caption{Classification results on MaxMin split. The best metric value in each group is marked in bold.}
\label{results_maxmin_split}
\end{table}

\begin{table}
\centering
\resizebox{\textwidth}{!}{
\begin{tabular}{|c|c|c|c|c|c|}
\hline
\textbf{Group} & \textbf{Method} & \textbf{MCC} & \textbf{AUROC} & \textbf{Precision} & \textbf{Recall} \\ \hline
\multirow{5}{*}{Fingerprints} & ECFP & \textbf{0.48 $\pm$ 0.02} & \textbf{78.16\% $\pm$ 1.20\%} & 72.95\% $\pm$ 2.23\% & 42.10\% $\pm$ 2.48\% \\ \cline{2-6} 
 & Layered & 0.46 $\pm$ 0.02 & 78.03\% $\pm$ 1.53\% & 73.67\% $\pm$ 2.13\% & 38.83\% $\pm$ 2.38\% \\ \cline{2-6} 
 & RDKit & 0.46 $\pm$ 0.02 & 75.44\% $\pm$ 1.46\% & \textbf{75.06\% $\pm$ 2.04\%} & 36.88\% $\pm$ 2.56\% \\ \cline{2-6} 
 & SECFP & 0.45 $\pm$ 0.03 & 75.75\% $\pm$ 1.03\% & 69.11\% $\pm$ 3.62\% & 41.56\% $\pm$ 3.16\% \\ \cline{2-6} 
 & \begin{tabular}[c]{@{}c@{}}Topological\\ Torsion\end{tabular} & 0.44 $\pm$ 0.02 & 75.09\% $\pm$ 1.14\% & 64.97\% $\pm$ 2.80\% & \textbf{43.46\% $\pm$ 1.93\%} \\ \hline
\multirow{3}{*}{Baselines} & Atom counts & 0.29 $\pm$ 0.04 & \textbf{75.15\% $\pm$ 0.88\%} & 46.10\% $\pm$ 3.31\% & \textbf{37.56\% $\pm$ 3.75\%} \\ \cline{2-6} 
 & LTP & 0.23 $\pm$ 0.01 & 70.66\% $\pm$ 0.88\% & 49.49\% $\pm$ 2.17\% & 22.15\% $\pm$ 0.66\% \\ \cline{2-6} 
 & MOLTOP & \textbf{0.33 $\pm$ 0.01} & 74.84\% $\pm$ 0.41\% & \textbf{58.00\% $\pm$ 0.97\%} & 30.34\% $\pm$ 1.21\% \\ \hline
\multirow{4}{*}{\begin{tabular}[c]{@{}c@{}}Graph\\ kernels\end{tabular}} & Propagation & 0.36 & 72.04\% & 51.43\% & \textbf{43.90\%} \\ \cline{2-6} 
 & Shortest paths & 0.31 & 71.03\% & 42.55\% & 48.78\% \\ \cline{2-6} 
 & WL & 0.41 & 70.50\% & 72.22\% & 31.71\% \\ \cline{2-6} 
 & WL-OA & \textbf{0.43} & \textbf{78.08\%} & \textbf{62.07\%} & \textbf{43.90\%} \\ \hline
\multirow{5}{*}{GNNs} & GCN & 0.30 $\pm$ 0.04 & 68.53\% $\pm$ 2.61\% & 50.50\% $\pm$ 7.07\% & 35.95\% $\pm$ 11.67\% \\ \cline{2-6} 
 & GraphSAGE & \textbf{0.33 $\pm$ 0.04} & \textbf{72.63\% $\pm$ 1.79\%} & \textbf{52.21\% $\pm$ 7.92\%} & 39.95\% $\pm$ 10.77\% \\ \cline{2-6} 
 & GIN & 0.32 $\pm$ 0.06 & 72.57\% $\pm$ 2.92\% & 43.22\% $\pm$ 8.64\% & \textbf{54.20\% $\pm$ 13.40\%} \\ \cline{2-6} 
 & GAT & 0.26 $\pm$ 0.05 & 68.35\% $\pm$ 2.84\% & 41.02\% $\pm$ 5.95\% & 40.10\% $\pm$ 7.37\% \\ \cline{2-6} 
 & AttentiveFP & 0.29 $\pm$ 0.06 & 70.76\% $\pm$ 2.27\% & 40.50\% $\pm$ 5.73\% & 50.49\% $\pm$ 6.04\% \\ \hline
\multirow{5}{*}{\begin{tabular}[c]{@{}c@{}}Pretrained\\ neural\\ networks\end{tabular}} & MAT & 0.25 & 63.88\% & 38.30\% & 43.90\% \\ \cline{2-6} 
 & R-MAT & 0.29 & \textbf{72.58\%} & 37.70\% & \textbf{56.10\%} \\ \cline{2-6} 
 & GROVER & 0.05 & 57.33\% & 22.58\% & 34.15\% \\ \cline{2-6} 
 & ChemBERTa & 0.27 & \textbf{72.58\%} & 35.29\% & 58.54\% \\ \cline{2-6} 
 & Mol2Vec & \textbf{0.31} & 69.10\% & \textbf{42.00\%} & 51.22\% \\ \hline
\end{tabular}
}
\caption{Classification results on time split. The best metric value in each group is marked in bold.}
\label{results_time_split}
\end{table}

\subsection{Comparison with MoleculeNet datasets}

\subsubsection{Molecular filters}

As a first method of datasets analysis, we utilize molecular filters, namely Lipinski, Ghose, Hao, Tice Insecticides, and Brenk. Table \ref{mol_filters} summarizes the results, where percentages indicate how many molecules from a given dataset pass a filter, that is, would be kept after applying it.

ApisTox has high values in the Lipinski and Ghose filters compared to other datasets, indicating that the properties of pesticides are similar to molecules with druglike potential. These properties may favor adsorption and bioavailability.

For Brenk filter, the results for ApisTox are similar to those obtained for the ClinTox and Tox21. A similar relationship is observed in the results of the Hao filter. This result suggests similarity, both physicochemical (Hao) and substructural (Brenk), between molecules toxic to humans and pesticides, which are by design toxic to other organisms, e.g. insects or weeds. The greatest differences between ApisTox, ClinTox and Tox21 collections appear in the results of the Tice Insecticides filter.

The observation that pesticides are relatively close to the chemical space of toxic drug-like molecules from medicinal chemistry also has practical implications, e.g. for substance repurposing. It could also be incorporated as a data augmentation strategy for pesticide ML models, helping the data scarcity problem in agrochemistry.

\begin{table}
\centering

\resizebox{\textwidth}{!}{
\begin{tabular}{|c|c|c|c|c|c|c|c|c|}
\hline
\textbf{Filter} & \textbf{ApisTox} & \textbf{BBBP} & \textbf{BACE} & \textbf{HIV} & \textbf{ClinTox} & \textbf{SIDER} & \textbf{ToxCast} & \textbf{Tox21} \\ \hline
Lipinski & 94.4 & 91.9 & 92.2 & 87.8 & 85.6 & 80.4 & 93.1 & 93.7 \\ \hline
Ghose & 60.9 & 61.2 & 28.5 & 59.9 & 45.6 & 43.8 & 46.4 & 49.8 \\ \hline
Hao & 70.9 & 67.4 & 25.8 & 51.8 & 51.5 & 46.8 & 70.2 & 71.4 \\ \hline
\begin{tabular}[c]{@{}c@{}}Tice\\ Insecticides\end{tabular} & 62.3 & 67.7 & 43.8 & 58.1 & 44.6 & 42.5 & 49.1 & 52.0 \\ \hline
Brenk & 41.0 & 62.0 & 78.3 & 34.6 & 51.1 & 50.5 & 42.0 & 43.7 \\ \hline
\end{tabular}
}
\caption{Percentage of molecules passing molecular filters.}
\label{mol_filters}
\end{table}

\subsubsection{Share of non-medical elements}

Some molecular representation methods operate on a small subset of elements deemed as ``medical'', instead of the full possible spectrum. They are often limited to only the most common atoms in the databases, while the others are treated as a single category, ``other elements''. This simplification can lead to an overreduction of the structural information of a molecule, which can consequently significantly affect the model results. Examples include MAT and R-MAT, yet, to our knowledge, this limitation has never been analyzed in other works employing those models.

\begin{figure}[H]
    \centering
    \includegraphics[width=0.8\textwidth]{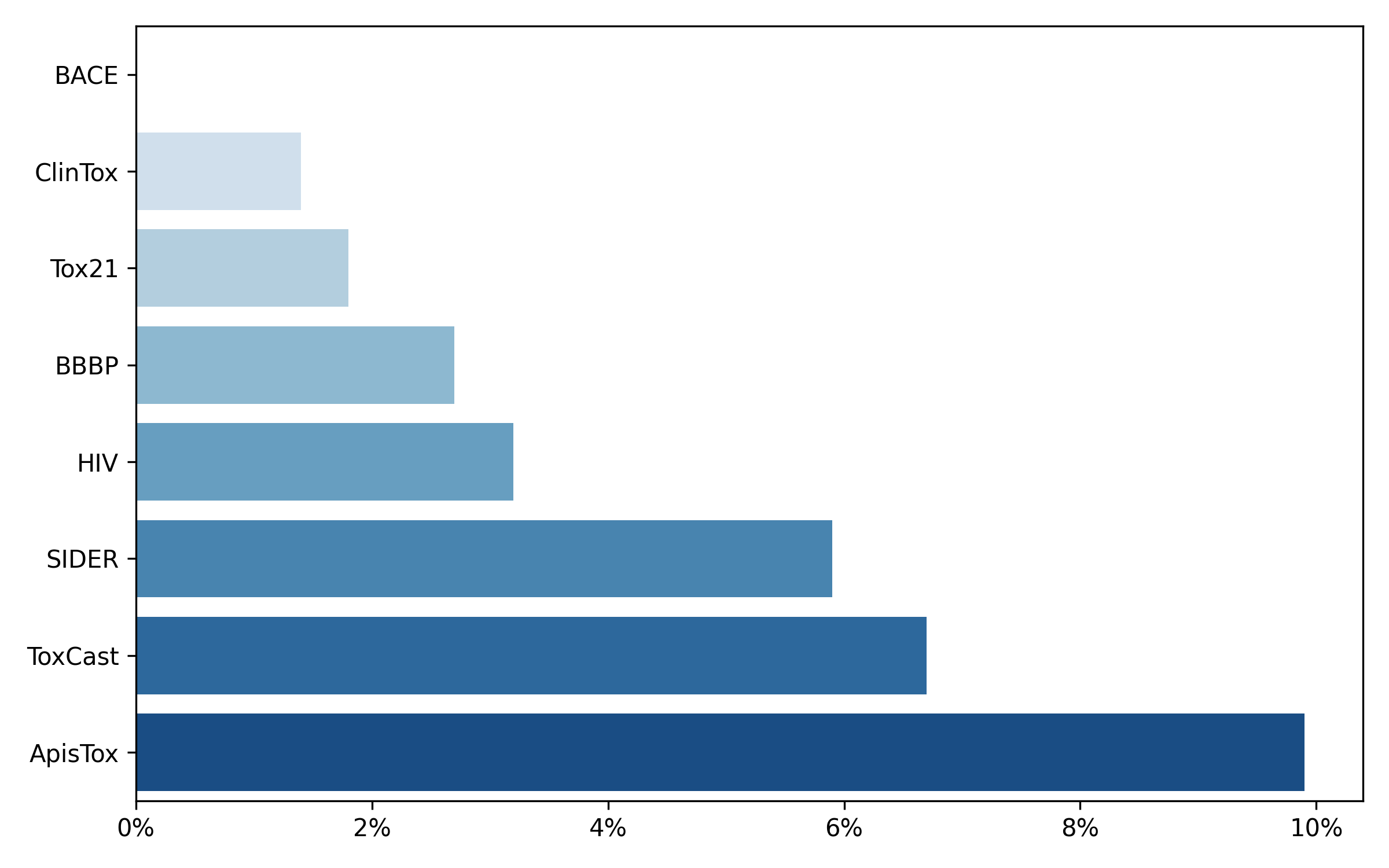}
    \caption{Percentage of molecules with ``non-medical'' atoms.}
    \label{non_med_atoms_fig}
\end{figure}

Following MAT and R-MAT papers, we consider the following elements as ``medical'': C, N, O, Si, Cl, S, F, P, B, Se, I, Br, As. One of the characteristic features of the ApisTox dataset is the significant proportion of molecules containing non-medical elements, as shown in Figure \ref{non_med_atoms_fig}.

The ApisTox dataset contains almost two times the number of such compounds compared to the MoleculeNet datasets. Overall, datasets that contain more toxic molecules on average, e.g. ToxCast and SIDER, tend to have a larger share of molecules with non-medical atoms. Note that MAT and R-MAT, which do not take such atoms into account, demonstrate only moderate performance on the ApisTox dataset.

\subsubsection{Share of fragmented molecules}

Fragmented molecules are defined as those in which two or more independent fragments are present in a single SMILES record. Examples include salts or mixtures, frequently found in agrochemical QSAR \cite{ApisTox}. Such compounds can be challenging for some methods, especially those based on graph neural networks (GNNs) or path-based fingerprints like Atom Pair. In such cases, the individual fragments are treated as separate graphs, which prevents the message passing between them. This can impact the performance of the models.

\begin{figure}[H]
    \centering
    \includegraphics[width=0.8\textwidth]{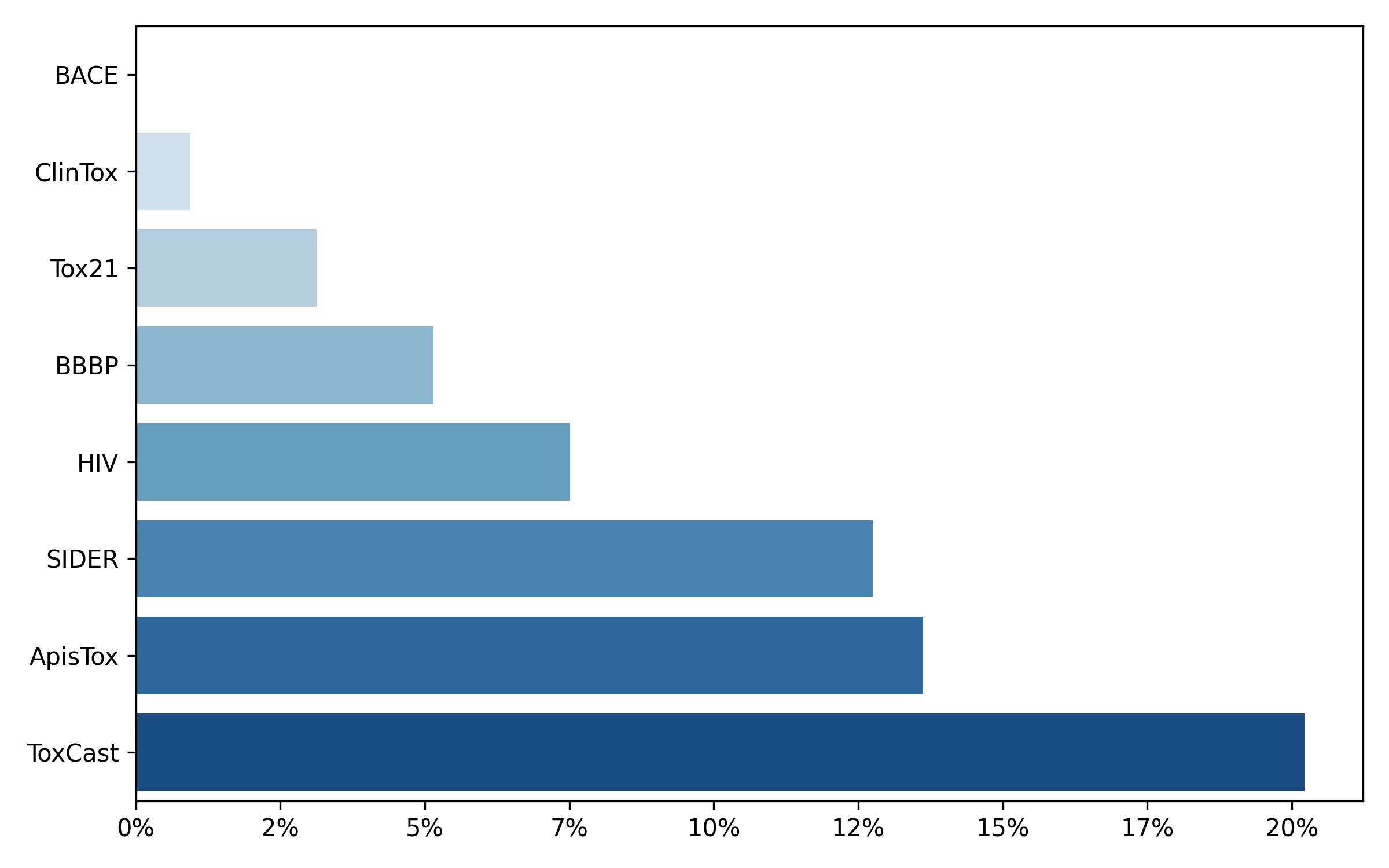}
    \caption{Percentage of fragmented molecules.}
    \label{frag_mols}
\end{figure}

We summarize the percentage of such compounds in Figure \ref{frag_mols}. ApisTox pesticides have a very high proportion of these molecules, compared to other data sets, almost 14\%, the second highest among the data sets analyzed. This may be one of the reasons why the GNNs achieve poor results on ApisTox.

\subsubsection{Unique functional groups}

Analysis of the distribution of functional groups allows us to determine the degree of uniqueness and structural variability of the datasets, based on the presence of distinct chemical structures. It also provides an intuitive approach to evaluating differences in the occupied chemical space, facilitating a comparative assessment of structural diversity among the datasets.

The functional groups defined in the Laggner fingerprint were utilized. It is a collection of 307 fragments, also including those typically indicating toxicity or generally less common in typical drug-like collections. The full list, including SMARTS definitions, is available at \url{https://github.com/scikit-fingerprints/scikit-fingerprints/blob/master/skfp/fingerprints/data/SMARTS_InteLigand.txt}.

\begin{figure}[H]
    \centering
    \includegraphics[width=0.8\textwidth]{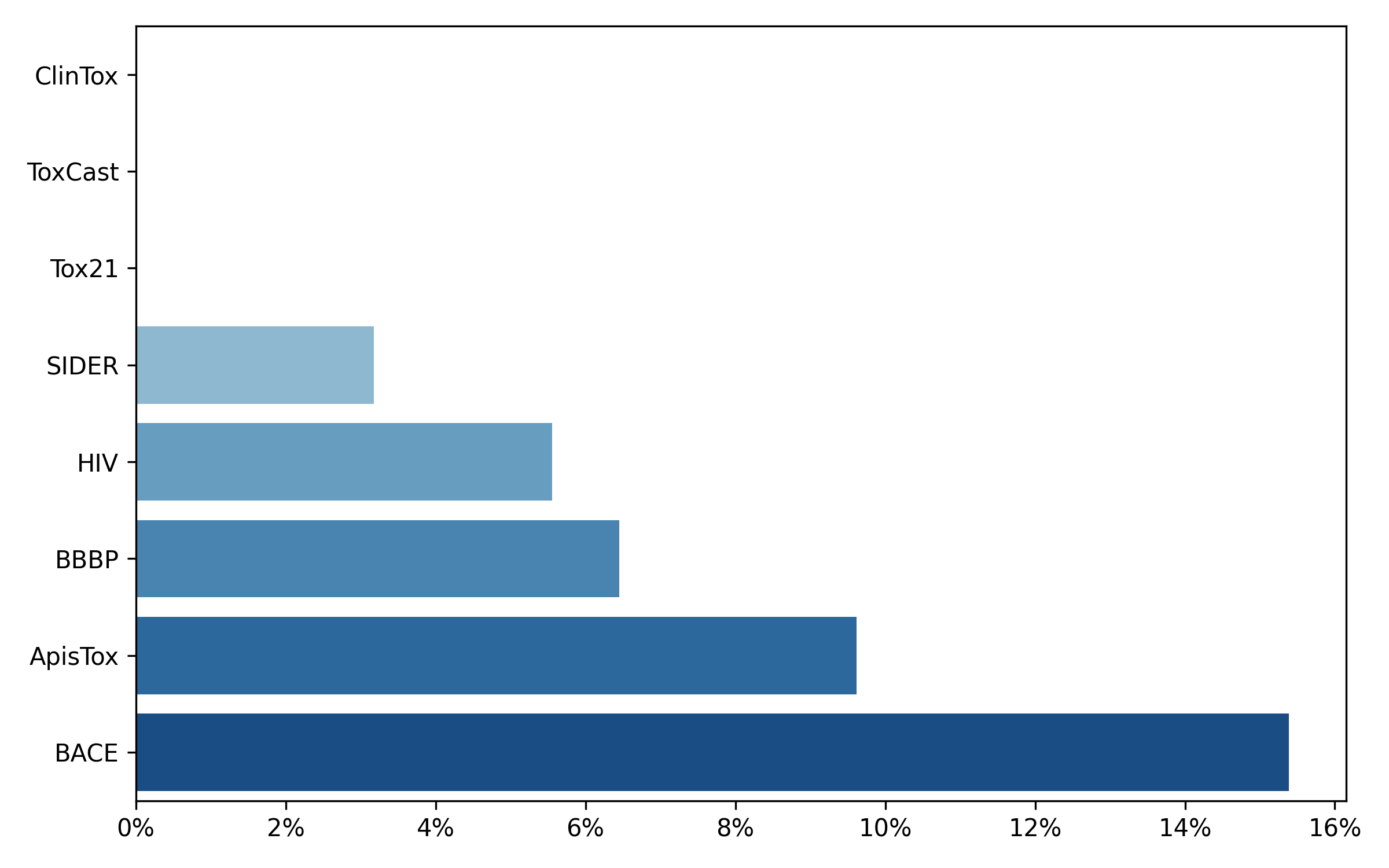}
    \caption{Percentage of unique SMARTS chemical fragments.}
    \label{unique_frag_fig}
\end{figure}

To remove outliers, we selected only groups that occur in at least 5\% of the molecules in a given dataset. As we want to quantify inter-dataset differences, we select only structures unique to each dataset, i.e., which occur in a given dataset and not in others. The results were normalized by dividing to the number of total fragments in the set and are shown in Figure \ref{unique_frag_fig}.

Analysis of the studied datasets indicates that the datasets concerning molecule toxicity, i.e. ClinTox, ToxCast, and Tox21, do not contain unique chemical fragments. Many of the fragments present in these datasets are shared with others, suggesting that the tested molecules occupy a similar chemical space in terms of functional groups. The BACE dataset contains the largest number of unique fragments, although its chemical space is the least diverse in terms of overall structures (see the next section). ApisTox demonstrates a relatively high uniqueness percentage, indicating that pesticides are structurally different in terms of functional groups. This is reasonable considering the different design goals of medicinal chemistry and agrochemistry.

\subsubsection{Chemical diversity}
\label{sec:chem_div}

ApisTox contains pesticides, while MoleculeNet focuses on drugs, drug candidates, and other drug-like substances. Those are potentially substantially different chemical spaces, which we quantify here. We use two methods: average pairwise Tanimoto similarity and the \#Circles measure.

\begin{figure}[H]
    \centering
    \includegraphics[width=0.8\textwidth]{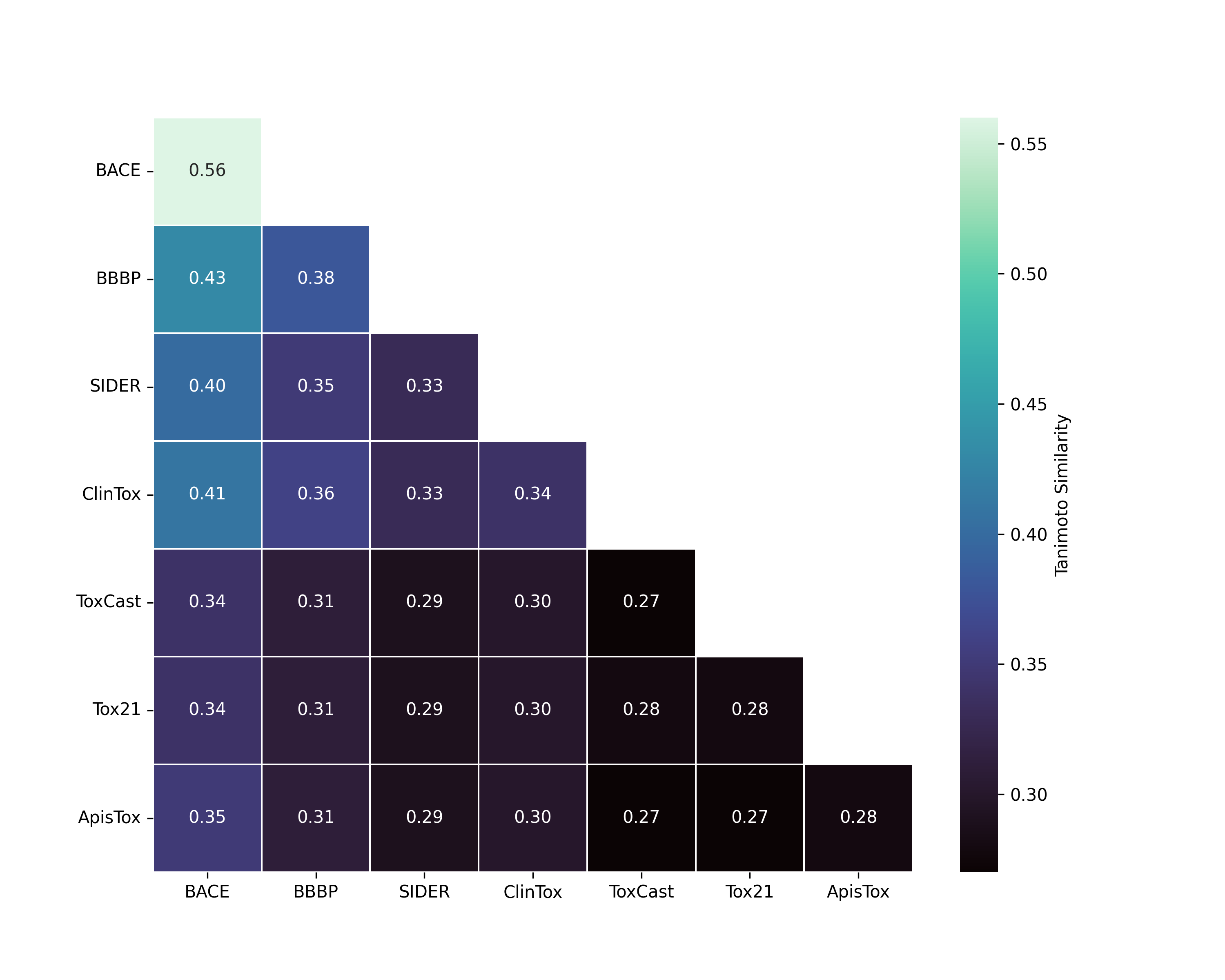}
    \caption{Average Tanimoto similarity between molecules from different datasets.}
    \label{tanimito_sim_map}
\end{figure}

Figure \ref{tanimito_sim_map} illustrates the average Tanimoto similarity calculated from PubChem fingerprints between pairs of molecules. Internal (intra-dataset) diversity is measured on diagonal, and off-diagonal entries show the external (inter-dataset) diversity, i.e. similarity between molecules from different datasets. PubChem fingerprints were used instead of ECFP4, because the latter resulted in near-identical distances for all datasets, even those obviously structurally different by manual inspection, e.g. BACE and ApisTox.

Datasets exhibiting the highest internal diversity, namely ToxCast, Tox21, and ApisTox, present comparable average similarity values, ranging from 0.275 to 0.284. It should be noted that the internal heterogeneity of a set shows a strong correlation with its heterogeneity in relation to other sets. This means that generally the greater the internal heterogeneity of a given set, the higher the external diversity values tend to be.

The BACE dataset has the lowest internal diversity, which can be attributed to its specific composition, which consists mainly of inhibitors of a single enzyme. This is in line with expectations based on chemical intuition.

\begin{figure}[H]
    \centering
    \includegraphics[width=0.8\textwidth]{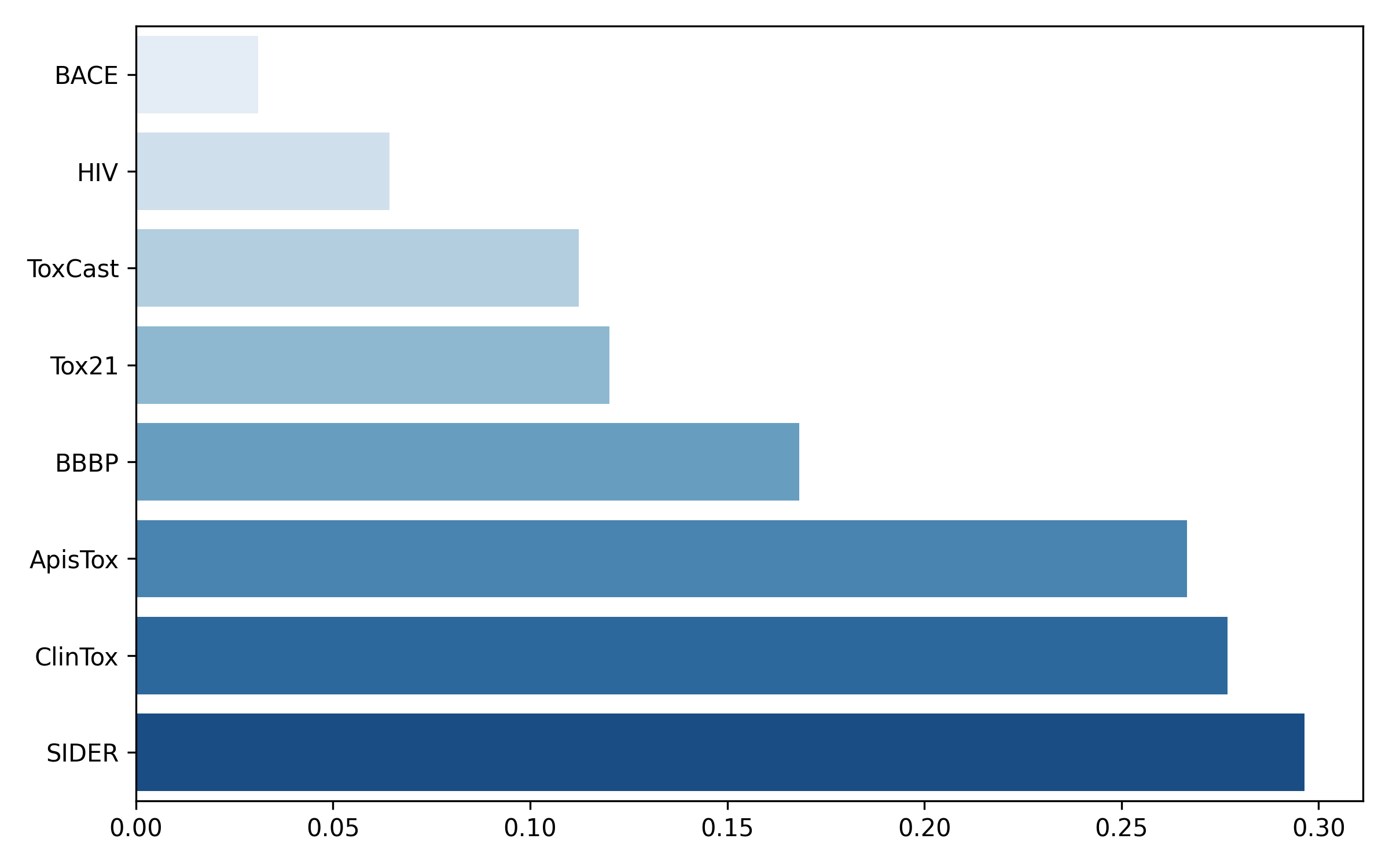}
    \caption{Normalized \#Circles measure values.}
    \label{n_circles_fig}
\end{figure}

Furthermore, all analyzed datasets appear to be relatively distant from each other (average Tanimoto similarity $\leq 0.56$), suggesting that they represent different areas of chemical space. However, it should be noted that the results obtained from this analysis show little correlation with the difficulty of their respective classification tasks and the quality of the resulting ML models \cite{MOLTOP}.

\#Circles measure is sensitive to the dataset size. To enable direct comparison, we normalize it, i.e. divide by the number of molecules in the given dataset. The results are shown in Figure \ref{n_circles_fig}.

The SIDER collection has the highest chemical diversity, while ApisTox and ClinTox rank just behind it. The high diversity of the chemical space poses a challenge for machine learning models, as it requires modeling a wide range of chemical compounds. In particular, it increases the chance of encountering novel substructures and functional groups, as well as activity cliffs, i.e. parts of chemical space where a small difference in structure significantly changes the compound properties.

\subsection{Interpretability - counterfactual explanations}

\begin{figure}[!ht]
  \centering
  \includegraphics[width=\textwidth]{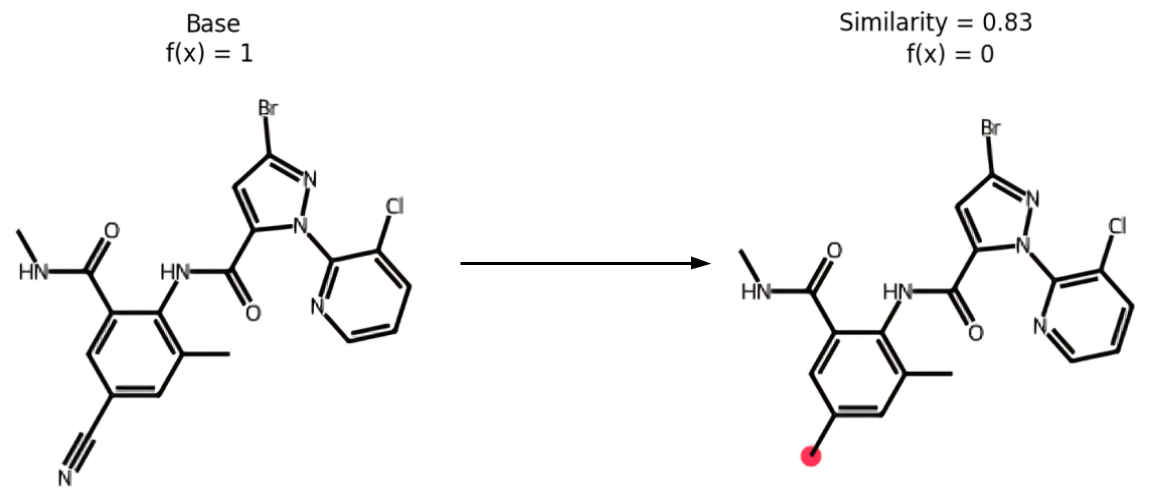}
  \caption{Example of a counterfactual perturbation for Cyantraniliprole, along with model predictions and ECFP4 Tanimoto similarity to the original compound.}
  \label{counterfactual_mol}
\end{figure}

We perform behavioral testing of trained models with counterfactual explanations. We use the MaxMin split here because it covers the entire chemical space well. The prediction model is SVM with the WL-OA kernel, as it gave the best results on this split. To avoid data leakage, we calculate counterfactual explanations only on test samples. In Figure \ref{counterfactual_mol}, we present an example of insecticide Cyantraniliprole, with all plots provided in the GitHub repository.

For example, when the nitrile group of Cyantraniliprole (an insecticide toxic to honeybees) is changed to methyl, its electronegativity and reactivity profile is modified. The rest of the structural features of the compound are preserved (e.g. benzamide ring system, pyrazole ring, and halogen substitutions), which may suggest that it could still bind its target ryanodine receptor. However, no release of cyanide or other reactive intermediates may also lead to a decrease in the toxicity to honey bee metabolism. The model changes its prediction from toxic to non-toxic accordingly. 

Overall, the most prevalent motifs influencing model predictions are connected to polarity, ionization states, charge distribution, and stability (e.g. ester bonds). In fact, these are some of the key factors that organic chemists and environmental scientists consider when designing molecules. The polarity and ionization states influence solubility and control the mobility of the compound in the environment, affecting aspects like runoff, leaching into groundwater, and bioavailability. More polar molecules tend to have lower bioaccumulation potential, as they interact less strongly with the lipid membranes of organisms \cite{pathak}. Charge characteristics can dictate how a molecule adsorbs in soils or sediments, affecting its overall persistence and exposure risks \cite{Bertram}. Finally, incorporating chemically labile groups like ester bonds is a common strategy to design molecules that break down into less harmful components after they have performed their intended function. This is part of a ``benign-by-design'' approach to reduce long-term environmental impact \cite{benign}.

\subsection{Interpretability - uncertain molecules}

In addition to the main dataset and its splits, ApisTox provides a smaller dataset of molecules rejected from the main dataset due to disagreements between experimental measurements in the ECOTOX database. For the main dataset, only molecules with perfectly agreeing $LD_{50}$ values were included, i.e., all under or over $11$ $\mu \text{g / bee}$ (threshold used in US EPA guidelines). Others are provided as a file with uncertain molecules. Performing predictions on those compounds provides additional evidence, in addition to the literature and experiments, about their toxicity. This is also a very practical application of proposed models, as conflicting measurements can easily arise in ecotoxicology research \cite{ApisTox}, e.g. due to variations in experimental procedures or outliers in results.

\begin{figure}[!ht]
  \centering
  \subfloat{\includegraphics[width=.48\textwidth]{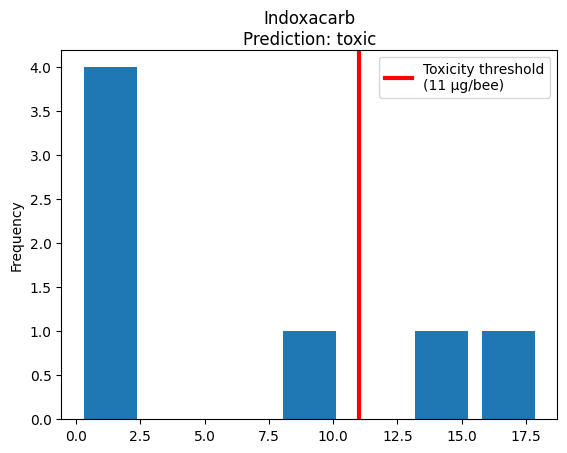}}\quad
  \subfloat{\includegraphics[width=.48\textwidth]{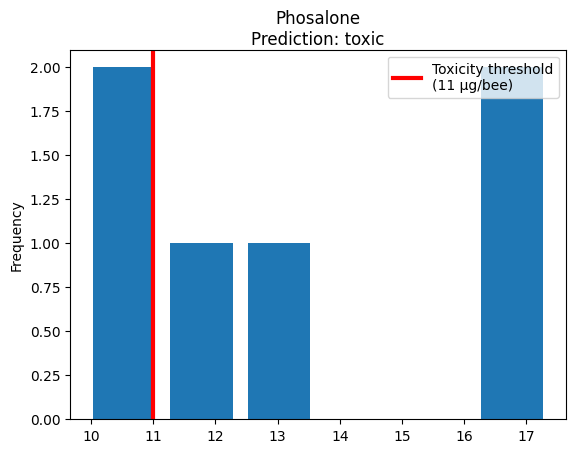}}\\
  \subfloat{\includegraphics[width=.48\textwidth]{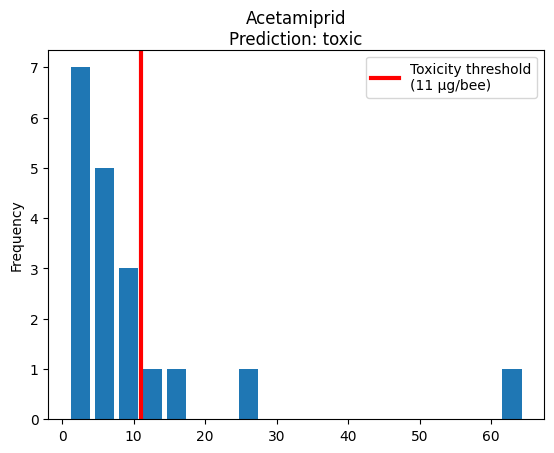}}\quad
  \subfloat{\includegraphics[width=.48\textwidth]{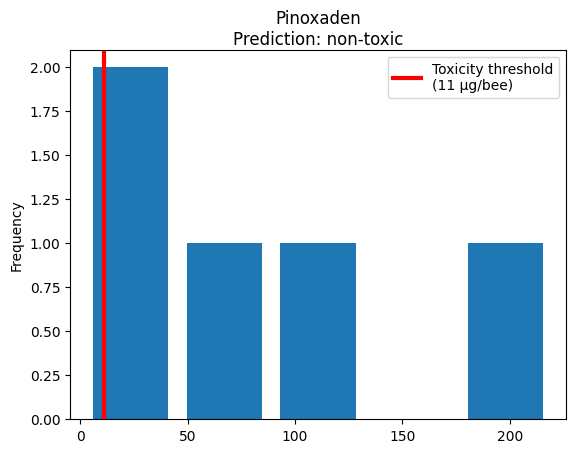}}
  \caption{Experimental measurements and model predictions for uncertain molecules.}
  \label{uncertain-mols-plots}
\end{figure}

We trained the SVM model with the WL-OA kernel on the entire ApisTox dataset and then made predictions on uncertain molecules. In Figure \ref{uncertain-mols-plots}, we present results for some of the most uncertain molecules, i.e., those that had the most disagreeing experimental measurements (below and above $11$ $\mu \text{g / bee}$). Each histogram shows the distribution of the experimental values for a given compound. All plots are included in the GitHub repository.

Indoxacarb, also known commercially as Steward, Avaunt, Advion, and Arilon, is an oxadiazine insecticide, used in many household pesticides, and is also approved for use in the US and UK. However, the EU withdrew its approval for the use of Indoxacarb as a plant protection insecticide at the end of 2021, due to its environmental impact and high toxicity. In particular, ``a high risk to bees was identified for the representative use in maize, sweet corn and lettuce for seed production'' \cite{eu_indoxacarb_decision}. Thus, this is a rather nuanced situation with regulatory disagreements, and prediction of machine learning further corroborates the high toxicity of Indoxacarb toward honey bees.

Phosalone has values ranging from $9.9$ to $17.4$ in the data. The model predicts toxicity (positive class), which is in agreement with the EU decision from 2006 \cite{eu_phosalone_decision}.

Acetamiprid has 15 out of 19 measurements that indicate toxicity, and our model agrees with this majority. However, its usage is universally allowed, including EU, US, and China. Predictions of ML models strengthen evidence supporting limitations of usage in such cases.

Lastly, Pinoxaden, a herbicide used against grass weeds, is typically recognized as having low toxicity for honey bees and is universally approved. 2 of 5 measurements in ECOTOX indicate toxicity, while our model predicts non-toxic label.

In conclusion, the created ML model generally agrees with the majority of experimental data, which is expected from a reasonable method. At the same time, it was also able to find more nuanced relations related to real-world regulatory status, which were not obvious from the data. Such applications of predictive modeling may be used as additional evidence for regulatory purposes.

\section{Conclusions}

This study provides a comprehensive evaluation of machine learning models to predict pesticide toxicity to honey bees using the ApisTox dataset. Findings highlight the distinct chemical space occupied by agrochemical compounds compared to medicinal datasets, emphasizing the need for domain-specific predictive models.

Among the evaluated methods, molecular fingerprints and graph kernels, particularly the WL-OA kernel, demonstrated superior performance. Somewhat surprisingly, more sophisticated models like graph neural networks (GNNs) and pretrained neural models struggled to generalize effectively. This suggests that current molecular embeddings may be strongly biased towards medicinal chemistry and may not generalize well to other chemical spaces like agrochemicals. Simpler models turned out to be more generalizable, both for MaxMin split and time split, which strengthens the observations of previous literature work like LTP and MOLTOP.

Analysis of the properties and diversity of ApisTox dataset revealed significant differences between pesticides and medicinal drug-like compounds, reinforcing the necessity of expanding benchmarks beyond the established medicinal chemistry applications.

The created models, on the example of SVM with WL-OA kernel, were further validated using interpretable machine learning techniques. Counterfactual explanations and tests on uncertain molecules show that they behave appropriately, focus on chemically relevant molecular features, and are able to understand nontrivial dependencies in ecotoxicology data. Those findings suggest that ML models can support regulatory assessments by providing additional predictions and evidence on toxicity classification.

In general, this research highlights the need for more expansive datasets and benchmarks for fair evaluation of molecular ML models. It also shows that established and simpler approaches are effective in predicting pesticide toxicity for honey bees. By advancing predictive ecotoxicology approaches, we can apply computer-aided drug design and rational drug design tools for the development of safer pesticides.

\section*{Acknowledgements}

We thank Wojciech Czech for support and useful discussions. In addition, we thank Piotr Ludynia, Michał Stefanik, and Adam Staniszewski for help with the development of scikit-fingerprints and the implementation of functionalities related to this paper. We also thank Alexandra Elbakyan for her work and support for accessibility of science.

The research presented in this paper was financed from funds assigned by the Polish Ministry of Science and Higher Education to the AGH University of Krakow and the Institute of Biochemistry and Biophysics of the Polish Academy of Sciences. Research project was supported by program ``Excellence initiative – research university'' for the AGH University of Krakow.

\bibliographystyle{elsarticle-num} 
\bibliography{bibliography}

\appendix

\section{Hyperparameters and model details}
\label{appendix_hyperparameters}

\subsection{Baselines}

\textbf{Atom counts} - no hyperparameters. Atomic numbers up to 89 (inclusive) were used. Classifier: Random Forest, 100 trees, \q{entropy} split criterion, \q{balanced} class weights. Using 100 trees rather than 500 gave better results during initial experiments.

\noindent\textbf{LTP} - no hyperparameters. Classifier follows original publication: Random Forest, 500 trees, \q{gini} split criterion, \q{balanced} class weights.

\noindent\textbf{MOLTOP} - no hyperparameters. Classifier follows original publication: Random Forest, 500 trees, \q{entropy} split criterion, minimum of 10 samples to split, \q{balanced} class weights.

\subsection{Fingerprints}

For all fingerprints, Random Forest was used as a classifier. Constant hyperparameters were: 100 trees, \q{entropy} split criterion, \q{balanced} class weights. \q{min\_samples\_split} was tuned in range [2, 10], alongside the hyperparameters for each fingerprint.

Fingerprints \textbf{Autocorrelation}, \textbf{Mordred} and \textbf{VSA} had no further hyperparameters. For VSA, the full space of possible features was used.

For substructural fingerprints \textbf{FunctionalGroups}, \textbf{GhoseCrippen}, \textbf{KlekotaRoth}, \textbf{Laggner}, \textbf{MACCS}, \textbf{MQNs}, \textbf{PubChem}, the only hyperparameter was whether to use binary or count variant, i.e. \q{count}: [False, True].

For \textbf{RDKit2DDescriptors} fingerprint, i.e. all RDKit descriptors combined, the only hyperparameter was whether to use CDF normalization proposed by D-MPNN \cite{DMPNN}.

Other fingerprints had hyperparameter grids as follows:

\textbf{Atom Pairs}: \q{fp\_size}: [512, 1024, 2048], \q{count}: [False, True]

\textbf{Avalon}: \q{fp\_size}: [256, 512, 1024, 2048], \q{count}: [False, True]

\textbf{ECFP}: \q{fp\_size}: [512, 1024, 2048], \q{radius}: [2, 3], \q{count}: [False, True]

\textbf{ERG}: \q{max\_path}: [5, 6, 7, \dots, 25]

\textbf{EState}: \q{variant}: [sum, bit, count]

\textbf{FCFP}: \q{fp\_size}: [512, 1024, 2048], \q{radius}: [2, 3], \q{count}: [False, True]

\textbf{Layered}: \q{fp\_size}: [512, 1024, 2048], \q{max\_path}: [5, 6, 7, 8, 9]

\textbf{Lingo}: \q{substring\_length}: [3, 4, 5, 6], \q{count}: [False, True]

\textbf{MAP}: \q{fp\_size}: [512, 1024, 2048], \q{radius}: [2, 3], \q{count}: [False, True]

\textbf{Pattern}: \q{fp\_size}: [512, 1024, 2048], \q{tautomers}: [False, True]

\textbf{PhysiochemicalProperties}: \q{fp\_size}: [512, 1024, 2048], \q{variant}: [BP, BT]

\textbf{RDKit}: \q{fp\_size}: [512, 1024, 2048], \q{max\_path}: [5, 6, 7, 8, 9], \q{count}: [False, True]

\textbf{SECFP}: \q{fp\_size}: [512, 1024, 2048], \q{radius}: [1, 2, 3, 4]

\textbf{TopologicalTorsion}: \q{fp\_size}: [512, 1024, 2048], \q{count}: [False, True]

\subsection{Graph kernels}

SVM used as a classifier for graph kernels always used \q{balanced} class weights. We always tuned the inverse regularization strength \q{C}: [1e-2, 1e-1, 1, 1e1, 1e2]. This was done in addition to kernel-specific hyperparameters, as listed below. Propagation, WL and WL-OA kernels were normalized. This was not possible for Shortest Path kernel due to disconnected graphs.

\textbf{Propagation}: \q{t\_max}: [1, 2, 3, 4, 5]

\textbf{Shortest paths}: no hyperparameters

\textbf{WL}: \q{n\_iter}: [1, 2, 3, 4, 5]

\textbf{WL-OA}: \q{n\_iter}: [1, 2, 3, 4, 5]

\subsection{Graph neural networks}

Hyperparameter grid for all GNNs was: \q{num\_layers}: [2, 3], \q{num\_channels}: [32, 64], \q{dropout}: [0.25, 0.5], \q{learning\_rate}: [1e-3, 1e-2, 1e-1]. Loss function with balanced class weights was always used (similarly to fingerprints models). Full batch training was used. Models were checked every 10 epochs and we selected the best model based on validation set results.

\subsection{Pretrained neural networks}

In all cases, logistic regression was used as a classifier, with 100 values of inverse regularization strength \q{C} checked on a logarithmic scale between 1e-4 and 1e4, as defined in \texttt{LogisticRegressionCV} in scikit-learn. There were no further hyperparameters to tune.

For \textbf{ChemBERTa}, we checked 3 variants available, based on amount of pretraining data:
\begin{itemize}
    \item small (\q{DeepChem/ChemBERTa-5M-MTR} on HuggingFace)
    \item medium (\q{DeepChem/ChemBERTa-10M-MTR})
    \item large (\mbox{\q{DeepChem/ChemBERTa-77M-MTR}})
\end{itemize}

For \textbf{GROVER}, we checked base and large versions, as defined by the original paper \cite{pretrained_GROVER}.

For \textbf{MAT}, we checked three size variants:
\begin{itemize}
    \item small (\q{mat\_masking\_200k})
    \item medium (\q{mat\_masking\_2M})
    \item large (\q{mat\_masking\_20M})
\end{itemize}

For \textbf{R-MAT}, we checked variants without and with additional RDKit descriptors.
\clearpage

\section{Full results of molecular fingerprints}
\label{appendix_fingerprints_full_results}

Here, we present full results of all molecular fingerprints.

\begin{table}[H]
\centering
\resizebox{\textwidth}{!}{
\begin{tabular}{|c|c|c|c|c|}
\hline
\textbf{Fingerprint}                                                & \textbf{MCC}    & \textbf{AUROC}       & \textbf{Precision}   & \textbf{Recall}      \\ \hline
AtomPairs                                                           & 0.37 $\pm$ 0.03 & 70.26\% $\pm$ 1.49\% & 63.38\% $\pm$ 3.19\% & 33.46\% $\pm$ 2.97\% \\ \hline
Avalon                                                              & 0.43 $\pm$ 0.02 & 77.28\% $\pm$ 1.02\% & 71.97\% $\pm$ 2.74\% & 35.37\% $\pm$ 2.03\% \\ \hline
Autocorrelation                                                     & 0.26 $\pm$ 0.05 & 70.43\% $\pm$ 1.38\% & 53.66\% $\pm$ 5.49\% & 23.76\% $\pm$ 4.02\% \\ \hline
ECFP                                                                & 0.48 $\pm$ 0.02 & 78.16\% $\pm$ 1.20\% & 72.95\% $\pm$ 2.23\% & 42.10\% $\pm$ 2.48\% \\ \hline
ERG                                                                 & 0.24 $\pm$ 0.04 & 64.98\% $\pm$ 1.37\% & 44.49\% $\pm$ 3.76\% & 30.15\% $\pm$ 3.04\% \\ \hline
EState                                                              & 0.28 $\pm$ 0.03 & 70.66\% $\pm$ 1.23\% & 49.45\% $\pm$ 3.11\% & 30.44\% $\pm$ 2.14\% \\ \hline
FCFP                                                                & 0.40 $\pm$ 0.03 & 76.13\% $\pm$ 0.98\% & 61.02\% $\pm$ 4.18\% & 40.20\% $\pm$ 3.21\% \\ \hline
FunctionalGroups                                                    & 0.31 $\pm$ 0.02 & 69.22\% $\pm$ 0.91\% & 44.50\% $\pm$ 2.20\% & 45.46\% $\pm$ 1.67\% \\ \hline
GhoseCrippen                                                        & 0.39 $\pm$ 0.02 & 72.45\% $\pm$ 1.07\% & 58.74\% $\pm$ 2.57\% & 40.83\% $\pm$ 1.60\% \\ \hline
KlekotaRoth                                                         & 0.36 $\pm$ 0.03 & 75.48\% $\pm$ 1.17\% & 58.03\% $\pm$ 3.42\% & 37.07\% $\pm$ 2.72\% \\ \hline
Laggner                                                             & 0.37 $\pm$ 0.03 & 76.72\% $\pm$ 1.00\% & 57.48\% $\pm$ 2.94\% & 38.63\% $\pm$ 2.90\% \\ \hline
Layered                                                             & 0.46 $\pm$ 0.02 & 78.03\% $\pm$ 1.53\% & 73.67\% $\pm$ 2.13\% & 38.83\% $\pm$ 2.38\% \\ \hline
Lingo                                                               & 0.32 $\pm$ 0.03 & 72.92\% $\pm$ 1.17\% & 57.17\% $\pm$ 4.76\% & 30.49\% $\pm$ 1.97\% \\ \hline
MACCS                                                               & 0.33 $\pm$ 0.02 & 74.46\% $\pm$ 1.36\% & 62.04\% $\pm$ 2.72\% & 27.61\% $\pm$ 1.33\% \\ \hline
MAP                                                                 & 0.31 $\pm$ 0.04 & 71.38\% $\pm$ 2.59\% & 75.04\% $\pm$ 6.47\% & 18.68\% $\pm$ 3.67\% \\ \hline
Mordred                                                             & 0.32 $\pm$ 0.03 & 71.84\% $\pm$ 1.57\% & 62.83\% $\pm$ 4.86\% & 26.00\% $\pm$ 2.21\% \\ \hline
MQNs                                                                & 0.39 $\pm$ 0.04 & 71.73\% $\pm$ 0.96\% & 61.91\% $\pm$ 2.95\% & 37.56\% $\pm$ 3.81\% \\ \hline
Pattern                                                             & 0.39 $\pm$ 0.03 & 74.74\% $\pm$ 1.29\% & 61.77\% $\pm$ 3.05\% & 38.10\% $\pm$ 3.41\% \\ \hline
\begin{tabular}[c]{@{}c@{}}Physiochemical\\ Properties\end{tabular} & 0.13 $\pm$ 0.04 & 60.14\% $\pm$ 1.06\% & 28.20\% $\pm$ 2.38\% & 40.68\% $\pm$ 3.50\% \\ \hline
PubChem                                                             & 0.39 $\pm$ 0.03 & 72.27\% $\pm$ 1.19\% & 65.93\% $\pm$ 3.82\% & 34.59\% $\pm$ 2.16\% \\ \hline
RDKit                                                               & 0.46 $\pm$ 0.02 & 75.44\% $\pm$ 1.46\% & 75.06\% $\pm$ 2.04\% & 36.88\% $\pm$ 2.56\% \\ \hline
RDKit2DDescriptors                                                  & 0.26 $\pm$ 0.04 & 72.84\% $\pm$ 1.51\% & 48.40\% $\pm$ 4.52\% & 27.90\% $\pm$ 4.20\% \\ \hline
SECFP                                                               & 0.45 $\pm$ 0.03 & 75.75\% $\pm$ 1.03\% & 69.11\% $\pm$ 3.62\% & 41.56\% $\pm$ 3.16\% \\ \hline
TopologicalTorsion                                                  & 0.44 $\pm$ 0.02 & 75.09\% $\pm$ 1.14\% & 64.97\% $\pm$ 2.80\% & 43.46\% $\pm$ 1.93\% \\ \hline
VSA                                                                 & 0.32 $\pm$ 0.05 & 70.43\% $\pm$ 1.26\% & 57.15\% $\pm$ 4.76\% & 30.88\% $\pm$ 4.04\% \\ \hline
\end{tabular}
}
\caption{Classification results on time split.}
\end{table}

\clearpage

\begin{table}[H]
\centering
\resizebox{\textwidth}{!}{
\begin{tabular}{|c|c|c|c|c|}
\hline
\textbf{Fingerprint}                                                & \textbf{MCC}    & \textbf{AUROC}       & \textbf{Precision}   & \textbf{Recall}      \\ \hline
AtomPairs                                                           & 0.45 $\pm$ 0.03 & 76.65\% $\pm$ 1.25\% & 70.16\% $\pm$ 3.63\% & 39.67\% $\pm$ 2.46\% \\ \hline
Avalon                                                              & 0.48 $\pm$ 0.03 & 76.09\% $\pm$ 1.27\% & 76.37\% $\pm$ 4.40\% & 39.67\% $\pm$ 1.48\% \\ \hline
Autocorrelation                                                     & 0.24 $\pm$ 0.04 & 71.51\% $\pm$ 1.39\% & 44.99\% $\pm$ 4.24\% & 29.19\% $\pm$ 3.36\% \\ \hline
ECFP                                                                & 0.42 $\pm$ 0.02 & 71.66\% $\pm$ 1.36\% & 66.49\% $\pm$ 2.83\% & 37.95\% $\pm$ 2.15\% \\ \hline
ERG                                                                 & 0.23 $\pm$ 0.03 & 67.73\% $\pm$ 1.01\% & 38.98\% $\pm$ 2.60\% & 38.95\% $\pm$ 1.95\% \\ \hline
EState                                                              & 0.38 $\pm$ 0.03 & 76.36\% $\pm$ 0.88\% & 55.50\% $\pm$ 2.66\% & 44.05\% $\pm$ 3.44\% \\ \hline
FCFP                                                                & 0.35 $\pm$ 0.02 & 75.92\% $\pm$ 1.06\% & 57.25\% $\pm$ 2.37\% & 36.33\% $\pm$ 2.37\% \\ \hline
FunctionalGroups                                                    & 0.39 $\pm$ 0.02 & 73.35\% $\pm$ 0.61\% & 42.94\% $\pm$ 1.20\% & 68.67\% $\pm$ 2.44\% \\ \hline
GhoseCrippen                                                        & 0.37 $\pm$ 0.02 & 74.69\% $\pm$ 1.19\% & 54.41\% $\pm$ 2.69\% & 42.62\% $\pm$ 1.73\% \\ \hline
KlekotaRoth                                                         & 0.42 $\pm$ 0.03 & 74.27\% $\pm$ 0.88\% & 57.85\% $\pm$ 3.66\% & 48.29\% $\pm$ 2.02\% \\ \hline
Laggner                                                             & 0.46 $\pm$ 0.03 & 77.45\% $\pm$ 0.84\% & 57.40\% $\pm$ 3.16\% & 56.95\% $\pm$ 3.01\% \\ \hline
Layered                                                             & 0.43 $\pm$ 0.02 & 77.00\% $\pm$ 1.26\% & 67.99\% $\pm$ 2.71\% & 39.33\% $\pm$ 2.25\% \\ \hline
Lingo                                                               & 0.35 $\pm$ 0.02 & 75.48\% $\pm$ 1.34\% & 61.61\% $\pm$ 3.21\% & 31.57\% $\pm$ 1.95\% \\ \hline
MACCS                                                               & 0.45 $\pm$ 0.03 & 79.77\% $\pm$ 0.96\% & 67.30\% $\pm$ 3.31\% & 42.62\% $\pm$ 3.10\% \\ \hline
MAP                                                                 & 0.34 $\pm$ 0.04 & 68.16\% $\pm$ 1.63\% & 69.02\% $\pm$ 5.04\% & 25.43\% $\pm$ 3.70\% \\ \hline
Mordred                                                             & 0.41 $\pm$ 0.04 & 78.33\% $\pm$ 1.41\% & 74.26\% $\pm$ 6.10\% & 31.38\% $\pm$ 2.80\% \\ \hline
MQNs                                                                & 0.35 $\pm$ 0.03 & 73.95\% $\pm$ 1.10\% & 57.64\% $\pm$ 3.80\% & 35.71\% $\pm$ 3.01\% \\ \hline
Pattern                                                             & 0.43 $\pm$ 0.03 & 77.84\% $\pm$ 0.81\% & 60.89\% $\pm$ 2.83\% & 45.86\% $\pm$ 2.81\% \\ \hline
\begin{tabular}[c]{@{}c@{}}Physiochemical\\ Properties\end{tabular} & 0.08 $\pm$ 0.03 & 59.27\% $\pm$ 0.78\% & 24.96\% $\pm$ 1.53\% & 39.71\% $\pm$ 3.38\% \\ \hline
PubChem                                                             & 0.44 $\pm$ 0.03 & 77.25\% $\pm$ 0.99\% & 61.32\% $\pm$ 3.65\% & 47.86\% $\pm$ 2.03\% \\ \hline
RDKit                                                               & 0.43 $\pm$ 0.03 & 74.52\% $\pm$ 1.28\% & 68.84\% $\pm$ 3.01\% & 37.81\% $\pm$ 2.36\% \\ \hline
RDKit2DDescriptors                                                  & 0.40 $\pm$ 0.04 & 77.87\% $\pm$ 1.16\% & 61.82\% $\pm$ 4.84\% & 39.57\% $\pm$ 3.33\% \\ \hline
SECFP                                                               & 0.41 $\pm$ 0.02 & 73.57\% $\pm$ 1.02\% & 64.81\% $\pm$ 2.20\% & 38.19\% $\pm$ 1.58\% \\ \hline
TopologicalTorsion                                                  & 0.41 $\pm$ 0.03 & 76.91\% $\pm$ 1.01\% & 62.06\% $\pm$ 3.39\% & 40.71\% $\pm$ 2.30\% \\ \hline
VSA                                                                 & 0.36 $\pm$ 0.03 & 73.70\% $\pm$ 1.30\% & 54.53\% $\pm$ 3.78\% & 41.19\% $\pm$ 2.53\% \\ \hline
\end{tabular}
}
\caption{Classification results on maxmin split.}
\end{table}

\clearpage

\section{Full results of pretrained neural models}
\label{appendix_pretrained_full_results}

Here, we present full results of all variants of pretrained neural models.

\begin{table}[H]
\centering
\resizebox{\textwidth}{!}{
\begin{tabular}{|c|c|c|c|c|}
\hline
\textbf{Fingerprint} & \textbf{MCC}     & \textbf{AUROC}       & \textbf{Precision}   & \textbf{Recall}      \\ \hline
ChemBERTa small      & 0.27 $\pm$ 0.00  & 72.58\% $\pm$ 0.00\% & 35.29\% $\pm$ 0.00\% & 58.54\% $\pm$ 0.00\% \\ \hline
ChemBERTa medium     & 0.24 $\pm$ 0.00  & 65.37\% $\pm$ 0.00\% & 35.09\% $\pm$ 0.00\% & 48.78\% $\pm$ 0.00\% \\ \hline
ChemBERTa large      & 0.24 $\pm$ 0.00  & 65.35\% $\pm$ 0.00\% & 35.00\% $\pm$ 0.00\% & 51.22\% $\pm$ 0.00\% \\ \hline
GROVER base          & 0.05 $\pm$ 0.00  & 57.33\% $\pm$ 0.00\% & 22.58\% $\pm$ 0.00\% & 34.15\% $\pm$ 0.00\% \\ \hline
GROVER large         & -0.02 $\pm$ 0.00 & 53.36\% $\pm$ 0.00\% & 18.67\% $\pm$ 0.00\% & 34.15\% $\pm$ 0.00\% \\ \hline
MAT small            & 0.25 $\pm$ 0.00  & 63.88\% $\pm$ 0.00\% & 38.30\% $\pm$ 0.00\% & 43.90\% $\pm$ 0.00\% \\ \hline
MAT medium           & 0.19 $\pm$ 0.00  & 66.90\% $\pm$ 0.00\% & 33.33\% $\pm$ 0.00\% & 41.46\% $\pm$ 0.00\% \\ \hline
MAT large            & 0.22 $\pm$ 0.00  & 61.18\% $\pm$ 0.00\% & 36.17\% $\pm$ 0.00\% & 41.46\% $\pm$ 0.00\% \\ \hline
R-MAT                & 0.29 $\pm$ 0.00  & 72.58\% $\pm$ 0.00\% & 37.70\% $\pm$ 0.00\% & 56.10\% $\pm$ 0.00\% \\ \hline
Mol2Vec              & 0.31 $\pm$ 0.00  & 69.10\% $\pm$ 0.00\% & 42.00\% $\pm$ 0.00\% & 51.22\% $\pm$ 0.00\% \\ \hline
\end{tabular}
}
\caption{Classification results on time split.}
\end{table}

\begin{table}[H]
\centering
\resizebox{\textwidth}{!}{
\begin{tabular}{|c|c|c|c|c|}
\hline
\textbf{Fingerprint} & \textbf{MCC} & \textbf{AUROC}    & \textbf{Precision} & \textbf{Recall}   \\ \hline
ChemBERTa small      & 0.25 $\pm$ 0.00 & 71.59\% $\pm$ 0.00\% & 34.29\% $\pm$ 0.00\%  & 57.14\% $\pm$ 0.00\% \\ \hline
ChemBERTa medium     & 0.37 $\pm$ 0.00 & 74.46\% $\pm$ 0.00\% & 42.86\% $\pm$ 0.00\%  & 64.29\% $\pm$ 0.00\% \\ \hline
ChemBERTa large      & 0.29 $\pm$ 0.00 & 71.53\% $\pm$ 0.00\% & 37.50\% $\pm$ 0.00\%  & 57.14\% $\pm$ 0.00\% \\ \hline
GROVER base          & 0.22 $\pm$ 0.00 & 71.46\% $\pm$ 0.00\% & 32.86\% $\pm$ 0.00\%  & 54.76\% $\pm$ 0.00\% \\ \hline
GROVER large         & 0.21 $\pm$ 0.00 & 66.98\% $\pm$ 0.00\% & 31.94\% $\pm$ 0.00\%  & 54.76\% $\pm$ 0.00\% \\ \hline
MAT small            & 0.36 $\pm$ 0.00 & 72.29\% $\pm$ 0.00\% & 40.58\% $\pm$ 0.00\%  & 66.67\% $\pm$ 0.00\% \\ \hline
MAT medium           & 0.18 $\pm$ 0.00 & 66.39\% $\pm$ 0.00\% & 30.14\% $\pm$ 0.00\%  & 52.38\% $\pm$ 0.00\% \\ \hline
MAT large            & 0.33 $\pm$ 0.00 & 73.17\% $\pm$ 0.00\% & 40.32\% $\pm$ 0.00\%  & 59.52\% $\pm$ 0.00\% \\ \hline
R-MAT                & 0.31 $\pm$ 0.00 & 70.46\% $\pm$ 0.00\% & 36.99\% $\pm$ 0.00\%  & 64.29\% $\pm$ 0.00\% \\ \hline
Mol2Vec              & 0.34 $\pm$ 0.00 & 76.57\% $\pm$ 0.00\% & 40.62\% $\pm$ 0.00\%  & 61.90\% $\pm$ 0.00\% \\ \hline
\end{tabular}
}
\caption{Classification results on maxmin split.}
\end{table}

\end{document}